\def\year{2019}\relax
\documentclass[letterpaper]{article} %DO NOT CHANGE THIS
\usepackage{aaai19}  %Required
\usepackage{times}  %Required
\usepackage{helvet}  %Required
\usepackage{courier}  %Required
\usepackage{url}  %Required
\usepackage{graphicx}  %Required
\usepackage{amssymb}
\usepackage{amsmath}
\usepackage{bm}
\usepackage{fancybox}
\usepackage{tikz}
\usetikzlibrary{positioning}
\usepackage[lofdepth,lotdepth]{subfig}
\usetikzlibrary{shapes,snakes}
\usepackage{todonotes}
\usepackage{adjustbox}
\usepackage{multirow}
\usepackage[ruled,vlined,linesnumbered]{algorithm2e}
\usepackage{algpseudocode}
\usepackage{bm,array}
\usepackage{colortbl}
\usepackage{tabularx}

\newcolumntype{C}{>{\centering\arraybackslash}p{4em}}
\newcolumntype{R}{>{\raggedright\arraybackslash}p{4em}}
\definecolor{Gray}{gray}{0.85}

\DeclareMathOperator{\argmax}{argmax}
\newtheorem{defi}{Definition}
\title{Improving Optimization Bounds Using Machine Learning:\\ Decision Diagrams Meet Deep Reinforcement Learning} 

\author{Quentin Cappart\textsuperscript{1,4}, Emmanuel Goutierre\textsuperscript{2},
David Bergman\textsuperscript{3} \and Louis-Martin Rousseau\textsuperscript{1} \\
  \textsuperscript{1}{Ecole Polytechnique de Montr\'{e}al, Montreal, Canada} \\
  \textsuperscript{2}{Ecole Polytechnique, Paris, France} \\
  \textsuperscript{3}{University of Connecticut, Stamford, CT 06901, USA} \\
  \textsuperscript{4}{Element AI, Montreal, Canada} \\
  \{quentin.cappart, louis-martin.rousseau\}@polymtl.ca \\
emmanuel.goutierre@polytechnique.edu \\
  david.bergman@uconn.edu}

\begin{document}

\maketitle

\begin{abstract}
Finding tight bounds on the optimal solution is a critical element of practical solution methods for discrete optimization problems.
In the last decade, decision diagrams (DDs) have brought a new perspective on obtaining upper and lower bounds that can be significantly better than classical bounding mechanisms, such as linear relaxations.   
It is well known that the quality of the bounds achieved through this flexible bounding method is highly reliant on the ordering of variables chosen for building the diagram, and finding an ordering that optimizes standard metrics is an NP-hard problem.
In this paper, we propose an innovative and generic approach based on deep reinforcement learning for obtaining an ordering for tightening the bounds obtained with relaxed and restricted DDs. We apply the approach to both the Maximum Independent Set Problem and the Maximum Cut Problem. Experimental results on synthetic instances show that the deep reinforcement learning approach, by achieving tighter objective function bounds,  generally outperforms ordering methods commonly used in the literature when the distribution of instances is known.  To the best knowledge of the authors, this is the first paper to apply machine learning to directly improve relaxation bounds obtained by general-purpose bounding mechanisms for combinatorial optimization problems.
\end{abstract}

\section{Introduction}
\label{sec:intro}

Relaxation bounds, and mechanisms by which those bounds can be improved, are perhaps the most critical element of scalable generic algorithms for discrete optimization problems.  As machine learning popularizes, a natural question arises: how can machine learning be used for improving optimization bounds ?  Finding a way to utilize the power of machine learning to prove tighter relaxation bounds may be a key for unlocking significant performance improvements in optimization solvers.
This paper provides, to the best  knowledge of the authors, a first effective approach in the literature towards achieving this goal.

The challenge one faces in using machine learning to tighten relaxation bounds is that the bound provided by classical methods (e.g., LP or SDP relaxations) are \emph{inflexible}; the algorithm used to solve the relaxation has no effect on the quality of the bound.
For example, given an IP model, the LP relaxation will report the same bound independent of what method is used to solve the relaxation and any other decision employed during the solution algorithm. 

Contrastingly, approximate decision diagrams (DDs) \cite{bergmanMDDManip}, a recently introduced optimization technology, provide a \emph{flexible} bounding method, in that decisions employed in the execution of the algorithms used to build the DDs directly affect the quality of the bound.
This is true for both relaxed DDs, that prove relaxation bounds, and restricted DDs, that identify primal solutions.   This opens the door for potential integration with machine learning. 

Initially introduced for representing switching circuits \cite{lee1959representation} and for formal verification \cite{bryant1986graph}, DDs in discrete optimization are used to encode the feasible solutions of a problem while preserving its combinatorial structure. 
A common application is to provide bounds, both upper and lower, for discrete optimization problems \cite{bergmanMDDManip,bergman2013optimization}.
However, the quality of the bounds is known to be tightly related 
to the variable ordering considered during the construction of the DD \cite{bergman2012variable}.  
It has been shown that finding an optimal ordering for general DDs is NP-hard  and is often challenging to even model. Besides, improving a given variable ordering is known to be NP-complete \cite{bollig1996improving}. Thus, designing methods for finding a good ordering is a hot topic in the community and continues as a challenge.  The idea suggested in this paper is to use machine learning to identify good variable orderings that therefore result in tighter objective function bounds.

%However, the size of a DD representing exactly a combinatorial problem may grow exponentially with the number of variables, which reduces its usability for large instances. Besides, the size of a DD is known to be tightly related to the variable ordering used for the construction. 

%Appropriate ordering can yield smaller DDs and better relaxation bounds but it has been shown that finding an optimal ordering for general DDs, or even improving a given variable ordering, is NP-complete \cite{bollig1996improving}. For such a reason, designing methods for finding a good ordering is a hot topic in the community and is still a challenge  \cite{bergman2012variable}.

In another field of research, reinforcement learning (RL) \cite{sutton1998reinforcement} is an area of machine learning focusing on how an agent can learn from its interactions with an environment. The agent moves from state to state by performing a sequence of actions, each of them giving a specific reward.  The behavior of an agent is characterized by a policy, determining which action should be taken from each state.
Given this context, the goal is to learn a policy maximizing the sum of rewards of each action done by the agent.

However, traditional  methods for RL suffer from a lack of scalability  and are limited to low-dimensional problems.
The main issue is that some  states are never considered during the learning process when large state spaces are considered.
Recently, deep learning \cite{lecun2015deep} provided new  tools  to overcome this problem.
The idea is to use a deep architecture as a function approximation for generalizing knowledge from visited to unknown states. 
Such an improvement enabled RL  to  scale  to  problems  that  were  previously  intractable. Notorious examples are the superhuman performances obtained for the game of Go \cite{silver2016mastering} and Atari 2600 \cite{mnih2013playing,mnih2015human}.
The combination of RL with a deep network is commonly referred as \textit{deep reinforcement learning} (DRL) \cite{arulkumaran2017brief}.  

Even more recently,  DRL has also been applied to identify high-quality primal bounds to some NP-hard combinatorial problems.
Most work focuses on the classical Traveling Salesman Problem \cite{bello2016neural,deudon2018learning}, with the exception of the approach of  Khalil et al. \cite{khalil2017learning} that tackles four NP-hard problems having a graph structure.
They use a deep learning architecture in order to embed the graph structure into features \cite{dai2016discriminative}.
The competitive results obtained suggest that this approach is a promising new tool for finding solutions to NP-hard problems. In this paper, we further push these efforts to be able to generate dual bounds.
%, especially when
%the problem structure is difficult to exploit, as the %variable ordering in decision diagrams. 

Given this related work, our contribution is positioned as follows. We propose a generic approach based on DRL in order to identify variable orderings for approximate DDs. The goal is to find orderings providing tight bounds. The focus is on relaxed DDs, as this provides a mechanism for utilizing machine learning to improve relaxation bounds, but we also show the effectiveness for restricted DDs, adding to the recent literature on using machine learning for finding high-quality heuristic solutions.  The approach has been validated on the Maximum Independent Set Problem, for which the variable ordering has been intensively studied \cite{bergman2012variable}. Its application to the Maximum Cut Problem is also considered.
We note that there has been limited work on applying machine learning to identify variable orderings for DDs in unrelated fields \cite{Car06,GruLivMar11}.  To the best of our knowledge, this work has not been extended to optimization. 

This paper is structured as follows. The next section introduces the technical background related to DDs and RL.
The process that we designed for learning an appropriate variable ordering is then presented. The RL model and the learning algorithms are detailed
and the construction of the DD using RL is described.
Finally, experiments on synthetic instances are carried out in the last section.

\section{Technical Background}
\label{sec:bg}

\subsection{Decision Diagrams}
In the optimization field, a \textit{decision diagram} (DD) 
is a structure that encodes a set of solutions to a constrained optimization problem (COP) 
$\langle X,D,C,O \rangle$ where $X$ is the set of variables,
$D$ the set of discrete domains restricting the values that variables
$x \in X$ can take, $C$ the sets of constraints and $O$ the objective function.
Formally, a DD associated to a combinatorial problem $P$
is a directed-layered-acyclic graph $B_P = \langle U,A,d \rangle$
where $U$ is the set of nodes, $A$ the set of arcs and $d$ is a function 
$A \to \mathbb{N}$ associating a label at each arc.
The set of nodes $U$ is partitioned into layers $L_i$, i.e., $U = \cup_{i=1}^m L_i$.
Layers $L_1$ and $L_m$ are both composed of a single
node: the root and the terminal node, respectively.
The width $w_i(B)$ of layer $L_i$ in a DD $B$ is defined
as the number of nodes in that layer: $w_i(B) = |L_i|$.
The width $w(B)$ of the DD is the maximum-width layer: 
$w(B) = \max_i w_i(B)$.
Each arc $a \in A$ is directed from a node in a layer $L_i$ to
a node in layer $L_{i+1}$  where $i\in[0,m-1]$.
The function $d$ associates to each arc $a$ a \emph{label} $d(a)$.
The arcs directed out of each node $u \in U$ have distinct labels, i.e., at most
one arc with tail $a$ having any domain value $d$.
We assume that for each $u$, there must exist a directed path from the
root node to $u$ and from $u$ to the terminal node.
A \emph{cost} $c(a) \in \mathbb{R}$ is also associated
to every arc in $a$, which is used to encode the objective function of solutions. 

In this paper, a DD $B_P$ for a COP $P$ has $n+1$ layers where 
$n$ is the number of decision variables in $P$. 
Each layer $L_i$ (except the last one) is linked to one variable $x_i$ of $P$ and  an arc $a$ 
from $L_i$ to $L_{i+1}$ with label $d(a)$ represents the assignment $x_i$ to $d(a)$.
A direct path from the root to the terminal node  of $B_P$ corresponds
then to a solution of $P$. 
The assignment of  variables in $P$ to layers during
the construction of the DD is referred as the \textit{variable ordering}.

A DD is \textit{exact} when the solutions encoded align exactly 
with the feasible solutions of the initial problem $P$ and for any arc-directed 
root-to-terminal node path $p$, 
the sum of the costs of the arcs equates to the evaluation 
of the objective function for the solution $x$ it corresponds, i.e., $\sum_{a \in p} c(a) = O(x)$. 
In this case, the longest path (assuming a maximization problem) from the root to the terminal node corresponds to the optimal solution of $P$.
However, the width of DDs tends to grow exponentially with the number of variables in the problem, which reduces its usability for large instances.
A DD is \textit{relaxed} when its encodes a superset of the feasible solutions of $P$ and for any arc-directed root-to-terminal node path $p$, the sum of the costs of the arcs is an upper bound (still in the case of a maximization problem) on the evaluation of the objective function for the solution $x$ it corresponds, i.e., $\sum_{a \in p} c(a) \geq O(x)$.
A relaxed DD can be  constructed incrementally, merging nodes on each layer until the width is below a threshold \cite{bergmanMDDManip} in such a way that no solution is lost during the merging process.
Hence, for a relaxed DD, the longest path gives an upper bound of the optimal solution for $P$. Finally, a DD is \textit{restricted} when it under-approximates the feasible solutions of $P$ and for any arc-directed root-to-terminal node path $p$,
the sum of the costs of the arcs is a lower bound on the evaluation of the objective function for the solution $x$ it corresponds, i.e., $\sum_{a \in p} c(a) \leq O(x)$.
There are several ways to construct such a DD, perhaps the simplest one is removing nodes from each layer once the width threshold is reached.  Unlike relaxed DDs, solutions are lost during the reduction.
For maximization problems, the longest path provides a lower bound of the optimal solution for $P$.
Optimization bounds can thereby be directly obtained through relaxed and restricted DDs.
Both take as input a specified maximum width, and it has been empirically shown that larger DDs generally provide tighter bounds but 
are in return more expensive to compute.
An exhaustive description of DDs and
their construction are provided in \cite{bergman2016decision}.

\subsection{Reinforcement Learning}

Let $\langle S,A,T,R \rangle$ be a tuple representing a deterministic couple agent-environment where $S$ is the set of states in the environment,
$A$  the set of actions that the agent can do, $T: S \times A \to S$ the transition function leading the agent from a state to another one given the action taken and $R : S \times A \to \mathbb{R}$ the reward function of taking an action from a specific state.
The behavior of an agent is defined by a policy $\pi : S \to A$, describing the action to be done given a specific state. 
The goal of an agent is to learn a policy maximizing the accumulated sum of rewards (eventually discounted) during its lifetime defined by a sequence of states $s_t \in S$ with $t \in [1,\Theta]$.  Such a sequence is called an \emph{episode} where $s_\Theta$ is the terminal state.
The expected return after time step $t$ is denoted by  $G_t  = \sum_{k=t+1}^{\Theta} \gamma^{k-t-1} R(s_k,a_k)$  where $\gamma \in [0,1]$ is a discounting factor used for parametrizing the weight of future rewards.  For a deterministic environment, The quality of  taking an action $a$ from a state $s$ under a policy $\pi$ is defined by the action-value function $Q^\pi(s,a) =  G_t$. The problem is to find a policy maximizing
the expected return: $\pi^\star = \argmax_{\pi} Q^\pi(s,a) ~ \forall s \in S, \forall a \in A$. In practice, $\pi^\star$ is computed from an initial policy and by two nested operations: (1) the policy evaluation, making the action-value function consistent with the current policy, and (2) the policy iteration, improving greedily the current policy.

However, in practice, the optimal policy, or even an optimal action-value function,  cannot be computed in a reasonable amount of time.
A method based on approximation, such as Q-learning \cite{watkins1992q}, is then required. Instead of computing the optimal action-value function,
Q-learning approximates the function by iteratively updating a current estimate after each action. 
%It is an off-policy method, which means that the update is independent of the policy being followed.
The update function is defined as follows: 
$Q(s_t,a_t) \xleftarrow{\alpha} R(s_t,a_t) + \gamma \max_{a \in A} Q(s_{t+1},a)$, where
$x \xleftarrow{\alpha} y$ denotes the update $x \leftarrow x + \alpha(y-x)$ and $\alpha \in(0,1]$ the learning rate.

Another issue arising for large problems is that almost every state encountered may never have been seen during previous updates, thus necessitating a method capable of utilizing prior knowledge to generalize for different states that share similarities. Among them,
neural fitted Q-learning \cite{riedmiller2005neural} uses a neural network for approximating the action-value function. 
This provides an estimator $\widehat{Q}(s,a,\textbf{w}) \approx Q(s,a)$ where $\textbf{w}$ is a weight vector that is learned. 
Stochastic Gradient Descent  \cite{bottou2010large} or another optimizer coupled with back-propagation \cite{rumelhart1986learning} is then used for updating $\textbf{w}$ and aims to minimize the squared loss 
between the current Q-value and the new value that should be assigned using Q-learning:
$\textbf{w} \leftarrow \textbf{w}- \frac{1}{2} \alpha \nabla L(\textbf{w})$ where the square loss is 
$L(\textbf{w}) = \big( R(s_t,a_t) + \gamma \max_{a \in A} \widehat{Q}(s_{t+1},a,\textbf{w})
- \widehat{Q}(s,a,\textbf{w}) \big)^2$.
Updates are done using \textit{experience replay}. Let $\langle s,a,r \rangle$ be a sample representing an action done at a specific state with its reward and $\mathcal{D}$ a sample store. Each time an action is performed, $\langle s,a,r \rangle$ is added in $\mathcal{D}$.
Then, the optimizer updates $\textbf{w}$ using a random sample taken from $\mathcal{D}$.

%\begin{align} 
%L(\textbf{w}) &= \Big( R(s_t,a_t) + \gamma \max_{a \in A} \widehat{Q}(s_{t+1},a,\textbf{w}) 
%- \widehat{Q}(s,a,\textbf{w}) \Big)^2 \notag \\ 
%\textbf{w} &\leftarrow \textbf{w}- \frac{1}{2} \alpha \nabla L(\textbf{w}) \label{eq:squaredloss}
%\end{align}

\section{Learning Process}
\label{sec:approach}

%\subsection{Maximum Independent Set Problem}
%
%Let $G(V,E)$ be a simple undirected graph.
%An independent set of $G$ is a subset of nodes $I\subseteq V$ such that there is no two vertices in $I$ that are connected by
%an edge of $E$. The maximum independent set problem consists in finding the independent set with the largest cardinality.

\subsection{Reinforcement Learning Formulation}
%Let us consider the Maximum Independent Set Problem (MISP) as a running example for describing our approach.

Designing a RL model for determining the variable ordering of a DD associated to the COP $\langle X,D,C,O \rangle$ requires  defining, adequately, the tuple $\langle S,A,T,R \rangle$ to represent the system. Our model is defined as follows.

\begin{description}
\item[State]  A state $s \in S$ is a pair $\langle s_L, s_B \rangle$ containing an ordered sequence of variables $s_L$ and a partially constructed  DD $s_B$
 associated with variables in $s_L$. A state $s$ is terminal if $s_L$ includes all the variables of $X$.

\item[Action]  An action  is defined as the selection of a variable from $X$. An action $a_s$ can be performed at state $s$ if and only if it is not yet inserted in $s_B$ ($a_s  \in X \setminus s_L$).

\item[Transition] A transition is a function updating a state according to the action performed. Let $ B \oplus x$ be an operator adding the variable $x$ into a decision diagram $B$ and $y :: x$ another operator appending the variable $x$ to the sequence $y$, we have $T(s,a_s) = \langle s_L :: a_s, s_B \oplus a_s \rangle$. 

\item[Reward] The reward function is designed to tighten the bounds obtained with the DD. 
When maximizing,  upper bounds are provided by relaxed DDs and lower bounds  by restricted DDs.
Both cases are associated with a common reward.   Let $\lceil B \rceil$ and $\lfloor B \rfloor$ indicate the current upper/lower bound obtained with the DD $B$. Such bounds correspond to the current partial longest path of the relaxed/restricted DD from the root node to the last constructed layer. At each variable insertion in $B$, the difference in the longest path when adding the new layer is computed. When computing the upper bound, this difference is penalized because we want the bound to be as small as possible: $R^{ub}(s,a_s) = - \big( \lceil s_B \rceil - \lceil s_B \oplus a_s \rceil \big)$. 
It is rewarded for the lower bound  where we want to increase it instead: 
$R^{lb} (s,a_s) =  \big( \lfloor s_B \rfloor  - \lfloor s_B \oplus a_s \rfloor \big)$. 
For minimization problems, the shortest path must be considered instead.
The upper bounds are then provided by restricted DDs and lower bounds by relaxed DDs.

%\begin{align} 
%R^{ub}(s,a_s) &= - \big( \lceil s_b \rceil - \lceil s_b \oplus a_s \rceil \big) \label{eq:ub} \\ 
%R^{lb} (s,a_s) &=  \big( \lfloor s_b \rfloor  - \lfloor s_b \oplus a_s \rfloor \big) \label{eq:lb}
%\end{align}

%Let us consider a maximization problem. Two situations are considered: the bound obtained with relax DDs, and the lower bound
%Let $\mathbb{B}$ be the set of possible decision diagrams and
% $\omega:  \mathbb{B} \to \mathbb{N}$ an operator giving the current width of a decision diagram, we have $R(s,a_s) = - \big( \omega(s_b) - \omega(s_b \oplus a_s) \big)$. In fact, this function defines a cost more than a reward. More generally, any unary operator $\ast: \mathbb{B} \to \mathbb{R}$ taking as input a decision diagram  can be used for the reward function.

\end{description}

Note that this formalization is generic and can be applied to any  problem that can be represented by a DD constructed layer-by-layer.
Indeed, all the problem-dependent characteristics are embedded into the DD construction and the insertion of variables (operator $\oplus$ in the transition function).  It is interesting to see that the construction of a DD can be elegantly formulated as a RL problem.
Indeed, both methods are based on dynamic programming and a recursive formulation which ease their integration.

\subsection{Learning Algorithm}

The basis of the learning algorithm relies on neural fitted Q-learning as described in the previous section and is presented in Algorithm~\ref{alg:learning}. At 
each iteration, a COP ($P$) is randomly taken from the training set and the learning is conducted on it. Effective learning for any particular class of COPs should consider instances for that class of COP.  For example, if the goal is to find objective function bounds for an instance of the  Maximum Independent Set Problem (formally defined later), other instances from that class of problem should be used during the training.
The algorithm returns a vector of weights ($\textbf{w}$) which is used for parametrizing the approximate action-value function $\widehat{Q}$.  
The basic algorithmic framework can be improved through the following: 
\begin{description}
\item[Mini-batches] Instead of updating the $\widehat{Q}$-function using a single sample as previously explained, it is also possible to update it by considering a mini-batch of $m$ samples from the store memory $\mathcal{D}$. As stressed by \cite{masters2018revisiting}, the choice of the mini-batch size can have a huge impact on the learning process. 
%On one hand, large batches results in faster learning and
% leverages better support for parallelism, especially on GPUs. On the other hand, small batches can provide a better generalization.
%  In our case, generalization is a critical concern and so are inclined to choose small mini-batches.
Let $L_j(\textbf{w})$ be the squared loss related to a sample $j$ with $N$ as the batch size; the gradient update, where the square loss of each sample is  summed, is a follows: $\textbf{w} \leftarrow \textbf{w} - \frac{1}{2N} \alpha  \sum_{j=1}^{N}  \nabla L_j(\textbf{w})$.

%\begin{equation}
%\label{eq:upd-grad}
%\textbf{w} \leftarrow \textbf{w} - \frac{1}{2B} \alpha  \sum_{j=1}^{B}  \nabla L_j(\textbf{w}).
%\end{equation} 

\item[Adaptive $\pmb{\epsilon}$-greedy] Always following a greedy policy results in a lack of exploration during learning. 
One solution is to introduce limited  randomness in choosing an action. 
 $\epsilon$-greedy refers to taking a random action with probability $\epsilon$ where $\epsilon \ll 1$. Otherwise, the current policy is followed.
 In our case, $\epsilon$ is adaptive and decreases linearly during the learning process, resulting in focused exploration at first followed by increasingly favoring exploitation.
% 
% \item[Adaptive learning rate] In order to stabilize the learning process and to guarantee convergence, the learning rate ($\alpha$) is exponentially decreased.  
%  
\item[Reward scaling] Gradient-based methods have difficulties to learn when  rewards are large or sparse.
\emph{Reward scaling} compresses the space of rewards into a smaller interval value near zero, while still remaining sufficiently large, since, as stressed by \cite{henderson2017deep}, tiny rewards can also lead to failed learning. 
We let $\rho \in \mathbb{R}$ be the scaling factor, generally defined as a power of 10, and rescale the rewards as $r_{t} = \rho R(s_t,a_t)$. 
\end{description}

\algrenewcomment[1]{\(\triangleright\) #1}

\begin{algorithm}[!ht]

\Comment{\textbf{Pre:} $\langle S,A,T,R \rangle$ is an environment tuple.}

\Comment{\hspace{0.7cm} $\mathcal{M}$ is the training set containing COPs.}

\Comment{\hspace{0.7cm} $K$ is the number of iterations.}

\Comment{\hspace{0.7cm} $N$ is the batch size.}

\Comment{\hspace{0.7cm} $\Theta$ is the length of an episode.}

\Comment{\hspace{0.7cm} $\epsilon, \rho, \alpha, \gamma$ are parameters as defined previously.}

\Comment{\hspace{0.7cm} $\pi, \textbf{w}$ are randomly initialized.}

$ $

$\mathcal{D} := \emptyset$ \hfill \Comment{\textit{Experience replay store}}

	\For{$i$ \textnormal{\textbf{from}} $1$ \textnormal{\textbf{to}} $K$}{
	
		$P :=  \mathrm{randomValueFrom}(\mathcal{M})$ 
		
		$\langle S,A,T,R \rangle := \mathrm{initializeEnvironment}(P)$ 
		
		$s_1 := \langle \emptyset , \emptyset \rangle $ \hfill \Comment{\textit{Decision diagram is empty}}
		 
			\For{$t$ \textnormal{\textbf{from}} $1$ \textnormal{\textbf{to}} $\Theta$}{

					$\pi := \argmax_{\pi} \widehat{Q}^\pi(s,a,\textbf{w}) \hfill \forall s \in S, \forall a \in A$
					
					$k :=  \mathrm{randomValueFrom}([0,1])$
							
					 \If{$k > \epsilon$} {
						 $a_t := \pi(s_t)$ \hfill \Comment{\textit{Following policy}}
					 }
					 
					 \Else {
					 	$a_t := \mathrm{randomValueFrom}(A)$ \hfill \Comment{\textit{ $\epsilon$-greedy}}
					 }

			$r_{t} := \rho R(s_t,a_t)  $ \hfill \Comment{\textit{Reward scaling}}
			
			$ s_{t+1} := T(s_t,a_t)$
										
			 $\mathcal{D} := \mathcal{D} \cup \{ \langle s_t, a_t, r_t  \rangle \} $ \hfill \Comment{\textit{Store update}}

			\For{$j$ \textnormal{\textbf{from}} $1$ \textnormal{\textbf{to}} $N$}{
				
				$e :=  \mathrm{randomValueFrom}(\mathcal{D})$
						
				$ L_j(\textbf{w})  :=   \mathrm{square ~ loss ~ using} ~ e ~\mathrm{and} ~ \gamma$
			}
					
			$\textbf{w} := \textbf{w} - \frac{1}{2N} \alpha  \sum_{j=1}^{N}  \nabla L_j(\textbf{w}) $ \hfill \Comment{\textit{Mini-batch}}
			
			$\mathrm{update}(\epsilon)$	
		}	
	
	}	
	
\Return $\textbf{w}$
\caption{Learning Algorithm.}

\label{alg:learning}
\end{algorithm}

\section{Decision Diagram Construction}
\label{sec:const}

Once the model has been trained, the next step is to use it in order to build the DD for a new instance. 
Let us illustrate the construction on the \textit{Maximum Independent Set Problem}.
\begin{defi}[Maximum Independent Set Problem]
\label{def:misp}
Let $G(V,E)$ be a simple undirected graph.
An independent set of $G$ is a subset of vertices $I\subseteq V$ such that there is no two vertices in $I$ that are connected by
an edge of $E$. The Maximum Independent Set Problem (MISP) consists in finding the independent set with the largest cardinality.
\end{defi}

Note that we use the term \textit{vertices} for elements of the graph and \textit{nodes} for DDs.
The problem is fully represented by a graph $G(V,E)$ and a classical formulation  assigns a binary variable $x_i$ for each vertex $i \in V$ indicating if the variable is selected  in the set or not. More details about the internal operations of the construction are provided in \cite{bergman2012variable}. 
Specific to the learning, the environment tuple is generated for $G$, and the learned model is then applied on it. The environment is directly infered from the previous formulation: the current state is  the list of variables $x_i$ already considered with the DD currently built, an action consists in choosing a new variable and the transition function with the reward is associated to the DD construction.  At each state, the model is called in order to compute the estimated $\widehat{Q}$-value for each action that can be performed in the current state. The network \textit{structureToVec} \cite{dai2016discriminative} can be used for parametrizing $\widehat{Q}$ \cite{khalil2017learning}.
The construction is  driven by the policy $\pi = \argmax_{a} \widehat{Q}(s,a)$. The best vertex according to the approximated action-value function is inserted in the DD at each step. 

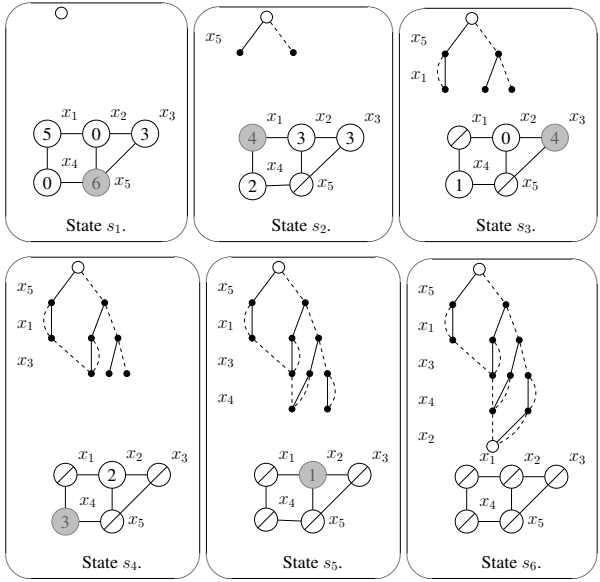
\begin{figure}[!ht]

\resizebox{0.95\columnwidth}{!}{
\subfloat{

\ovalbox{

\begin{tikzpicture}[
level/.style={sibling distance=15mm/#1, level distance=10mm},
every label/.append style={font=\LARGE},
every node/.style={font=\LARGE}
]
\node [circle,draw] (z){};

\node [below left = 0.2cm and 1cm  of z] (v1){};
\node [below = 5mm of v1] (v2){};

\node[circle, draw, below left = 3cm and 0.cm of z, label={above right:$x_1$}] (m0) {5};
\node[circle, draw, right = 0.6cm of m0, label={above right:$x_2$}] (m1) {0}; 
\node[circle, draw, right = 0.6cm of m1, label={above right:$x_3$}] (m2) {3}; 
\node[circle, draw, below = 0.6cm of m0, label={above right:$x_4$}] (m3) {0}; 
\node[circle, draw, below = 0.6cm of m1, label={right:$x_5$}, fill=gray, opacity=0.5] (m4) {6}; 

\path (m0) edge[-] node {} (m1);
\path (m0) edge[-] node {} (m3);
\path (m1) edge[-] node {} (m2);
\path (m1) edge[-] node {} (m4);
\path (m3) edge[-] node {} (m4);
\path (m2) edge[-] node {} (m4);

\node [below = 0.5cm of m4] (leg){State $s_1$.};

\end{tikzpicture} 
}

}

\subfloat{
\ovalbox{

\begin{tikzpicture}[
level/.style={sibling distance=15mm/#1, level distance=10mm},
every label/.append style={font=\LARGE},
every node/.style={font=\LARGE}
]
\node [circle,draw] (z){}
  child[solid] {node  [circle,draw,solid,fill,scale=0.5] (n0) {}
  }
  child[dashed] {node [circle,draw,solid,fill,scale=0.5] (n2)  {}
};

\node [below left = 0.2cm and 1cm  of z] (v1){$x_5$};
\node [below = 5mm of v1] (v2){};

\node[circle, draw, below left = 3cm and 0.cm of z, label={above right:$x_1$}, fill=gray, opacity=0.5] (m0) {4};
\node[circle, draw, right = 0.6cm of m0, label={above right:$x_2$}] (m1) {3}; 
\node[circle, draw, right = 0.6cm of m1, label={above right:$x_3$}] (m2) {3}; 
\node[circle, draw, below = 0.6cm of m0, label={above right:$x_4$}] (m3) {2}; 
\node[forbidden sign, ,minimum size=6.5mm, draw, below = 0.6cm of m1, label={right:$x_5$}] (m4) {}; 

\path (m0) edge[-] node {} (m1);
\path (m0) edge[-] node {} (m3);
\path (m1) edge[-] node {} (m2);
\path (m1) edge[-] node {} (m4);
\path (m3) edge[-] node {} (m4);
\path (m2) edge[-] node {} (m4);

\node [below = 0.5cm of m4] (leg){State $s_2$.};
\end{tikzpicture} 
}
}
\subfloat{
\ovalbox{

\begin{tikzpicture}[
level/.style={sibling distance=15mm/#1, level distance=10mm},
every label/.append style={font=\LARGE},
every node/.style={font=\LARGE}
]

\node [circle,draw] (z){}
  child[solid] {node  [circle,draw,solid,fill,scale=0.5] (n0) {}
    child[solid] {node  [circle,draw,solid,fill,scale=0.5] (n1) {}
%      child {node {$\vdots$}
%        child {node [circle,draw] (d) {$\frac{n}{2^k}$}}
%        child {node [circle,draw] (e) {$\frac{n}{2^k}$}}
%     } 
%      child {node {$\vdots$}}
    }
  }
  child[dashed] {node [circle,draw,solid,fill,scale=0.5] (n2)  {}
    child[solid] {node  [circle,draw,solid,fill,scale=0.5] (n3) {}
    }
  child[dashed] {node  [circle,draw,solid,fill,scale=0.5] (l) {}
%    child {node {$\vdots$}}
%    child {node (c){$\vdots$}
      %child {node [circle,draw] (o) {$\frac{n}{2^k}$}}
%      child {node [circle,draw] (p) {$\frac{n}{2^k}$}
%        child [grow=right] {node (q) {$=$} edge from parent[draw=none]
%          child [grow=right] {node (q) {$O_{k = \lg n}(n)$} edge from parent[draw=none]
%            child [grow=up] {node (r) {$\vdots$} edge from parent[draw=none]
%              child [grow=up] {node (s) {$O_2(n)$} edge from parent[draw=none]
%                child [grow=up] {node (t) {$O_1(n)$} edge from parent[draw=none]
%                  child [grow=up] {node (u) {$O_0(n)$} edge from parent[draw=none]}
%                }
%              }
%            }
%            child [grow=down] {node (v) {$O(n \cdot \lg n)$}edge from parent[draw=none]}
%          }
%        }
%      }
%    }
  }
};
 \path (n0)  edge[dashed,bend right=40] node {} (n1);
%\path (b) -- (g) node [midway] {+};
%\path (k) -- (l) node [midway] {+};
%\path (k) -- (g) node [midway] {+};
%\path (d) -- (e) node [midway] {+};
%\path (o) -- (p) node [midway] {+};
%\path (o) -- (e) node (x) [midway] {$\cdots$}
%  child [grow=down] {
%    node (y) {$O\left(\displaystyle\sum_{i = 0}^k 2^i \cdot \frac{n}{2^i}\right)$}
%    edge from parent[draw=none]
%  };
%\path (q) -- (r) node [midway] {+};
%\path (s) -- (r) node [midway] {+};
%\path (s) -- (t) node [midway] {+};
%\path (s) -- (l) node [midway] {=};
%\path (t) -- (u) node [midway] {+};
%\path (z) -- (u) node [midway] {=};
%\path (j) -- (t) node [midway] {=};
%\path (y) -- (x) node [midway] {$\Downarrow$};
%\path (v) -- (y)
%  node (w) [midway] {$O\left(\displaystyle\sum_{i = 0}^k n\right) = O(k \cdot n)$};
%\path (q) -- (v) node [midway] {=};
%\path (e) -- (x) node [midway] {+};
%\path (o) -- (x) node [midway] {+};
%\path (y) -- (w) node [midway] {$=$};
%\path (v) -- (w) node [midway] {$\Leftrightarrow$};
%\path (r) -- (c) node [midway] {$\cdots$};

\node [below left = 0.2cm and 1cm  of z] (v1){$x_5$};
\node [below = 5mm of v1] (v2){$x_1$};

\node[forbidden sign, ,minimum size=6.5mm, draw, below left = 3cm and 0.cm of z, label={above right:$x_1$}] (m0) {};
\node[circle, draw, right = 0.6cm of m0, label={above right:$x_2$}] (m1) {0}; 
\node[circle, draw, right = 0.6cm of m1, label={above right:$x_3$}, fill=gray, opacity=0.5] (m2) {4}; 
\node[circle, draw, below = 0.6cm of m0, label={above right:$x_4$}] (m3) {1}; 
\node[forbidden sign, ,minimum size=6.5mm, draw, below = 0.6cm of m1, label={right:$x_5$}] (m4) {}; 

\path (m0) edge[-] node {} (m1);
\path (m0) edge[-] node {} (m3);
\path (m1) edge[-] node {} (m2);
\path (m1) edge[-] node {} (m4);
\path (m3) edge[-] node {} (m4);
\path (m2) edge[-] node {} (m4);

\node [below = 0.5cm of m4] (leg){State $s_3$.};

\end{tikzpicture} 

}

}
}

\resizebox{0.95\columnwidth}{!}{
\subfloat{

\ovalbox{

\begin{tikzpicture}[
level/.style={sibling distance=15mm/#1, level distance=10mm},
every label/.append style={font=\LARGE},
every node/.style={font=\LARGE}
]

\node [circle,draw] (z){}
  child[solid] {node  [circle,draw,solid,fill,scale=0.5] (n0) {}
    child[solid] {node  [circle,draw,solid,fill,scale=0.5] (n1) {}
%      child {node {$\vdots$}
%        child {node [circle,draw] (d) {$\frac{n}{2^k}$}}
%        child {node [circle,draw] (e) {$\frac{n}{2^k}$}}
%     } 
%      child {node {$\vdots$}}
    }
  }
  child[dashed] {node [circle,draw,solid,fill,scale=0.5] (n2)  {}
    child[solid] {node  [circle,draw,solid,fill,scale=0.5] (n3) {}
      child[solid] {node  [circle,draw,solid,fill,scale=0.5] (n4) {}
      }
    }
  child[dashed] {node  [circle,draw,solid,fill,scale=0.5] (l) {}
      child[solid] {node  [circle,draw,solid,fill,scale=0.5] (n5) {}}
      child[dashed] {node  [circle,draw,solid,fill,scale=0.5] (n6) {}
      }
%    child {node {$\vdots$}}
%    child {node (c){$\vdots$}
      %child {node [circle,draw] (o) {$\frac{n}{2^k}$}}
%      child {node [circle,draw] (p) {$\frac{n}{2^k}$}
%        child [grow=right] {node (q) {$=$} edge from parent[draw=none]
%          child [grow=right] {node (q) {$O_{k = \lg n}(n)$} edge from parent[draw=none]
%            child [grow=up] {node (r) {$\vdots$} edge from parent[draw=none]
%              child [grow=up] {node (s) {$O_2(n)$} edge from parent[draw=none]
%                child [grow=up] {node (t) {$O_1(n)$} edge from parent[draw=none]
%                  child [grow=up] {node (u) {$O_0(n)$} edge from parent[draw=none]}
%                }
%              }
%            }
%            child [grow=down] {node (v) {$O(n \cdot \lg n)$}edge from parent[draw=none]}
%          }
%        }
%      }
%    }
  }
};
 \path (n0)  edge[dashed,bend right=40] node {} (n1);
 \path (n3)  edge[dashed,bend left=40] node {} (n4);
\path (n1) edge[dashed] node {} (n4);
%\path (b) -- (g) node [midway] {+};
%\path (k) -- (l) node [midway] {+};
%\path (k) -- (g) node [midway] {+};
%\path (d) -- (e) node [midway] {+};
%\path (o) -- (p) node [midway] {+};
%\path (o) -- (e) node (x) [midway] {$\cdots$}
%  child [grow=down] {
%    node (y) {$O\left(\displaystyle\sum_{i = 0}^k 2^i \cdot \frac{n}{2^i}\right)$}
%    edge from parent[draw=none]
%  };
%\path (q) -- (r) node [midway] {+};
%\path (s) -- (r) node [midway] {+};
%\path (s) -- (t) node [midway] {+};
%\path (s) -- (l) node [midway] {=};
%\path (t) -- (u) node [midway] {+};
%\path (z) -- (u) node [midway] {=};
%\path (j) -- (t) node [midway] {=};
%\path (y) -- (x) node [midway] {$\Downarrow$};
%\path (v) -- (y)
%  node (w) [midway] {$O\left(\displaystyle\sum_{i = 0}^k n\right) = O(k \cdot n)$};
%\path (q) -- (v) node [midway] {=};
%\path (e) -- (x) node [midway] {+};
%\path (o) -- (x) node [midway] {+};
%\path (y) -- (w) node [midway] {$=$};
%\path (v) -- (w) node [midway] {$\Leftrightarrow$};
%\path (r) -- (c) node [midway] {$\cdots$};

\node [below left = 0.2cm and 1cm  of z] (v1){$x_5$};
\node [below = 5mm of v1] (v2){$x_1$};
\node [below = 5mm of v2] (v3){$x_3$};

\node[forbidden sign, ,minimum size=6.5mm, draw, below left = 5.5cm and 0.cm of z, label={above right:$x_1$}] (m0) {};
\node[circle, draw, right = 0.6cm of m0, label={above right:$x_2$}] (m1) {2}; 
\node[forbidden sign, ,minimum size=6.5mm, draw, right = 0.6cm of m1, label={above right:$x_3$}] (m2) {}; 
\node[circle, draw, below = 0.6cm of m0, label={above right:$x_4$}, fill=gray, opacity=0.5] (m3) {3}; 
\node[forbidden sign, ,minimum size=6.5mm, draw, below = 0.6cm of m1, label={right:$x_5$}] (m4) {}; 

\path (m0) edge[-] node {} (m1);
\path (m0) edge[-] node {} (m3);
\path (m1) edge[-] node {} (m2);
\path (m1) edge[-] node {} (m4);
\path (m3) edge[-] node {} (m4);
\path (m2) edge[-] node {} (m4);

\node [below = 0.5cm of m4] (leg){State $s_4$.};

\end{tikzpicture} 

}

}

\subfloat{
\ovalbox{

\begin{tikzpicture}[
level/.style={sibling distance=15mm/#1, level distance=10mm},
every label/.append style={font=\LARGE},
every node/.style={font=\LARGE}
]

\node [circle,draw] (z){}
  child[solid] {node  [circle,draw,solid,fill,scale=0.5] (n0) {}
    child[solid] {node  [circle,draw,solid,fill,scale=0.5] (n1) {}
%      child {node {$\vdots$}
%        child {node [circle,draw] (d) {$\frac{n}{2^k}$}}
%        child {node [circle,draw] (e) {$\frac{n}{2^k}$}}
%     } 
%      child {node {$\vdots$}}
    }
  }
  child[dashed] {node [circle,draw,solid,fill,scale=0.5] (n2)  {}
    child[solid] {node  [circle,draw,solid,fill,scale=0.5] (n3) {}
      child[solid] {node  [circle,draw,solid,fill,scale=0.5] (n4) {}
      	child[dashed] {node  [circle,draw,solid,fill,scale=0.5] (n8) {}
     	 }
      }
    }
  child[dashed] {node  [circle,draw,solid,fill,scale=0.5] (l) {}
      child[solid] {node  [circle,draw,solid,fill,scale=0.5] (n5) {}}
      child[dashed] {node  [circle,draw,solid,fill,scale=0.5] (n6) {}
      	child[solid] {node [circle,draw,solid,fill,scale=0.5] (n7) {}}
      }
%    child {node {$\vdots$}}
%    child {node (c){$\vdots$}
      %child {node [circle,draw] (o) {$\frac{n}{2^k}$}}
%      child {node [circle,draw] (p) {$\frac{n}{2^k}$}
%        child [grow=right] {node (q) {$=$} edge from parent[draw=none]
%          child [grow=right] {node (q) {$O_{k = \lg n}(n)$} edge from parent[draw=none]
%            child [grow=up] {node (r) {$\vdots$} edge from parent[draw=none]
%              child [grow=up] {node (s) {$O_2(n)$} edge from parent[draw=none]
%                child [grow=up] {node (t) {$O_1(n)$} edge from parent[draw=none]
%                  child [grow=up] {node (u) {$O_0(n)$} edge from parent[draw=none]}
%                }
%              }
%            }
%            child [grow=down] {node (v) {$O(n \cdot \lg n)$}edge from parent[draw=none]}
%          }
%        }
%      }
%    }
  }
};
 \path (n0)  edge[dashed,bend right=40] node {} (n1);
 \path (n3)  edge[dashed,bend left=40] node {} (n4);
 \path (n5)  edge[dashed,bend left=20] node {} (n8);
\path (n1) edge[dashed] node {} (n4);
\path (n5) edge[-] node {} (n8);
\path (n6) edge[dashed, bend left=40] node {} (n7);
%\path (b) -- (g) node [midway] {+};
%\path (k) -- (l) node [midway] {+};
%\path (k) -- (g) node [midway] {+};
%\path (d) -- (e) node [midway] {+};
%\path (o) -- (p) node [midway] {+};
%\path (o) -- (e) node (x) [midway] {$\cdots$}
%  child [grow=down] {
%    node (y) {$O\left(\displaystyle\sum_{i = 0}^k 2^i \cdot \frac{n}{2^i}\right)$}
%    edge from parent[draw=none]
%  };
%\path (q) -- (r) node [midway] {+};
%\path (s) -- (r) node [midway] {+};
%\path (s) -- (t) node [midway] {+};
%\path (s) -- (l) node [midway] {=};
%\path (t) -- (u) node [midway] {+};
%\path (z) -- (u) node [midway] {=};
%\path (j) -- (t) node [midway] {=};
%\path (y) -- (x) node [midway] {$\Downarrow$};
%\path (v) -- (y)
%  node (w) [midway] {$O\left(\displaystyle\sum_{i = 0}^k n\right) = O(k \cdot n)$};
%\path (q) -- (v) node [midway] {=};
%\path (e) -- (x) node [midway] {+};
%\path (o) -- (x) node [midway] {+};
%\path (y) -- (w) node [midway] {$=$};
%\path (v) -- (w) node [midway] {$\Leftrightarrow$};
%\path (r) -- (c) node [midway] {$\cdots$};

\node [below left = 0.2cm and 1cm  of z] (v1){$x_5$};
\node [below = 5mm of v1] (v2){$x_1$};
\node [below = 5mm of v2] (v3){$x_3$};
\node [below = 5mm of v3] (v4){$x_4$};

\node[forbidden sign, ,minimum size=6.5mm, draw, below left = 5.5cm and 0.cm of z, label={above right:$x_1$}] (m0) {};
\node[circle, draw, right = 0.6cm of m0, label={above right:$x_2$}, fill=gray, opacity=0.5] (m1) {1}; 
\node[forbidden sign, ,minimum size=6.5mm,draw, right = 0.6cm of m1, label={above right:$x_3$}] (m2) {}; 
\node[forbidden sign, ,minimum size=6.5mm, draw, below = 0.6cm of m0, label={above right:$x_4$}] (m3) {}; 
\node[forbidden sign, ,minimum size=6.5mm, draw, below = 0.6cm of m1, label={right:$x_5$}] (m4) {}; 

\path (m0) edge[-] node {} (m1);
\path (m0) edge[-] node {} (m3);
\path (m1) edge[-] node {} (m2);
\path (m1) edge[-] node {} (m4);
\path (m3) edge[-] node {} (m4);
\path (m2) edge[-] node {} (m4);

\node [below = 0.5cm of m4] (leg){State $s_5$.};

\end{tikzpicture} 

}
}
\subfloat{
\ovalbox{

\begin{tikzpicture}[
level/.style={sibling distance=15mm/#1, level distance=10mm},
every label/.append style={font=\LARGE},
every node/.style={font=\LARGE}
]

\node [circle,draw] (z){}
  child[solid] {node  [circle,draw,solid,fill,scale=0.5] (n0) {}
    child[solid] {node  [circle,draw,solid,fill,scale=0.5] (n1) {}
%      child {node {$\vdots$}
%        child {node [circle,draw] (d) {$\frac{n}{2^k}$}}
%        child {node [circle,draw] (e) {$\frac{n}{2^k}$}}
%     } 
%      child {node {$\vdots$}}
    }
  }
  child[dashed] {node [circle,draw,solid,fill,scale=0.5] (n2)  {}
    child[solid] {node  [circle,draw,solid,fill,scale=0.5] (n3) {}
      child[solid] {node  [circle,draw,solid,fill,scale=0.5] (n4) {}
      	child[dashed] {node  [circle,draw,solid,fill,scale=0.5] (n8) {}
      		child[dashed] {node  [circle,draw,solid] (n9) {}}
     	 }
      }
    }
  child[dashed] {node  [circle,draw,solid,fill,scale=0.5] (l) {}
      child[solid] {node  [circle,draw,solid,fill,scale=0.5] (n5) {}}
      child[dashed] {node  [circle,draw,solid,fill,scale=0.5] (n6) {}
      	child[solid] {node [circle,draw,solid,fill,scale=0.5] (n7) {}}
      }
%    child {node {$\vdots$}}
%    child {node (c){$\vdots$}
      %child {node [circle,draw] (o) {$\frac{n}{2^k}$}}
%      child {node [circle,draw] (p) {$\frac{n}{2^k}$}
%        child [grow=right] {node (q) {$=$} edge from parent[draw=none]
%          child [grow=right] {node (q) {$O_{k = \lg n}(n)$} edge from parent[draw=none]
%            child [grow=up] {node (r) {$\vdots$} edge from parent[draw=none]
%              child [grow=up] {node (s) {$O_2(n)$} edge from parent[draw=none]
%                child [grow=up] {node (t) {$O_1(n)$} edge from parent[draw=none]
%                  child [grow=up] {node (u) {$O_0(n)$} edge from parent[draw=none]}
%                }
%              }
%            }
%            child [grow=down] {node (v) {$O(n \cdot \lg n)$}edge from parent[draw=none]}
%          }
%        }
%      }
%    }
  }
};
 \path (n0)  edge[dashed,bend right=40] node {} (n1);
 \path (n3)  edge[dashed,bend left=40] node {} (n4);
 \path (n5)  edge[dashed,bend left=20] node {} (n8);
  \path (n7)  edge[dashed,bend left=20] node {} (n9);
\path (n1) edge[dashed] node {} (n4);
\path (n5) edge[-] node {} (n8);
\path (n6) edge[dashed,bend left=40] node {} (n7);
\path (n7) edge[-] node {} (n9);
%\path (b) -- (g) node [midway] {+};
%\path (k) -- (l) node [midway] {+};
%\path (k) -- (g) node [midway] {+};
%\path (d) -- (e) node [midway] {+};
%\path (o) -- (p) node [midway] {+};
%\path (o) -- (e) node (x) [midway] {$\cdots$}
%  child [grow=down] {
%    node (y) {$O\left(\displaystyle\sum_{i = 0}^k 2^i \cdot \frac{n}{2^i}\right)$}
%    edge from parent[draw=none]
%  };
%\path (q) -- (r) node [midway] {+};
%\path (s) -- (r) node [midway] {+};
%\path (s) -- (t) node [midway] {+};
%\path (s) -- (l) node [midway] {=};
%\path (t) -- (u) node [midway] {+};
%\path (z) -- (u) node [midway] {=};
%\path (j) -- (t) node [midway] {=};
%\path (y) -- (x) node [midway] {$\Downarrow$};
%\path (v) -- (y)
%  node (w) [midway] {$O\left(\displaystyle\sum_{i = 0}^k n\right) = O(k \cdot n)$};
%\path (q) -- (v) node [midway] {=};
%\path (e) -- (x) node [midway] {+};
%\path (o) -- (x) node [midway] {+};
%\path (y) -- (w) node [midway] {$=$};
%\path (v) -- (w) node [midway] {$\Leftrightarrow$};
%\path (r) -- (c) node [midway] {$\cdots$};

\node [below left = 0.2cm and 1cm  of z] (v1){$x_5$};
\node [below = 5mm of v1] (v2){$x_1$};
\node [below = 5mm of v2] (v3){$x_3$};
\node [below = 5mm of v3] (v4){$x_4$};
\node [below = 5mm of v4] (v5){$x_2$};

\node[forbidden sign, ,minimum size=6.5mm, draw, below left = 5.5cm and 0.cm of z, label={above right:$x_1$}] (m0) {};
\node[forbidden sign, ,minimum size=6.5mm, draw, right = 0.6cm of m0, label={above right:$x_2$}] (m1) {}; 
\node[forbidden sign, ,minimum size=6.5mm, draw, right = 0.6cm of m1, label={above right:$x_3$}] (m2) {}; 
\node[forbidden sign, ,minimum size=6.5mm, draw, below = 0.6cm of m0, label={above right:$x_4$}] (m3) {}; 
\node[forbidden sign, ,minimum size=6.5mm, draw, below = 0.6cm of m1, label={right:$x_5$}] (m4) {}; 

\path (m0) edge[-] node {} (m1);
\path (m0) edge[-] node {} (m3);
\path (m1) edge[-] node {} (m2);
\path (m1) edge[-] node {} (m4);
\path (m3) edge[-] node {} (m4);
\path (m2) edge[-] node {} (m4);

\node [below = 0.5cm of m4] (leg){State $s_6$.};

\end{tikzpicture} 

}

}
}

\caption{Example of an exact DD construction for a MISP instance, following policy $\pi = \argmax_{a} Q^\pi(s,a)$. }
\label{fig:dd-const}
\end{figure}

This process is illustrated in Figure \ref{fig:dd-const} for a MISP instance. Solid arcs indicate that the vertex related to the current variable is selected in the solution while dashed arcs indicate the opposite.
The partially constructed DD and the inserted/remaining vertices are depicted for each state.
The value in each vertex indicates the $\widehat{Q}$-value computed by the model for each  state-action pair. Gray vertices are the ones that are greedily selected by the policy. No vertices can be inserted twice. The construction is terminated when all vertices are inserted.

\section{Experimental Results}
\label{sec:expe}

Our first set of experiments are  carried out on the MISP, for which  the impact of variable ordering has been deeply studied \cite{bergman2013optimization}. The last experiments analyze the generalization of the approach on the Maximum Cut Problem. For the MISP, the approach is compared with the linear programming relaxation bound, random orderings,  and three ordering heuristics commonly used in the literature:

\begin{enumerate}

\item \textit{Linear Programming Relaxation \textnormal{(\texttt{LP})}}: The value of the linear relaxation obtained using a standard clique formulation for the MISP as described in  \cite{bergman2013optimization}. 
%Let us notice that this relaxation can only be compared with relaxed DDs as they both provide feasible solutions

\item \textit{Random Selection \textnormal{(\texttt{RAND})}}:  An ordering of the vertices is drawn uniformly at random from all permutations. For each test, 100 random trials are performed and the average, best and worst results are reported.

\item \textit{Maximal Path Decomposition \textnormal{(\texttt{MPD})}}:  A \emph{maximal path decomposition} is precomputed and used as the ordering of the vertices \cite{bergman2013optimization}. This  ordering bounds the width of the exact DDs by the Fibonacci numbers.

\item \textit{Minimum Number of States \textnormal{(\texttt{MIN})}}: Having constructed up to layer $j$ and hence chosen the first $j-1$ vertices, the next vertex is selected as the one appearing in the fewest number of
states in the DD nodes in layer $j$. This heuristic aims to minimize greedily the size of the subsequent layer.

\item \textit{Minimum Vertex Degree \textnormal{(\texttt{DEG})}}: The vertices are ordered in ascending order of vertex degree. 
The vertices with the lowest degree are inserted first.

\end{enumerate}

%\subsection{Datasets Used}

%\begin{description}
%\item[Synthetic dataset]
%Synthetic instances are generated either following the Erdos-Rendy \cite{erdos1960evolution} or the Barabasi-Albert \cite{albert2002statistical} model, which are standard representations of many real-world graphs. Erdos-Rendy gaphs are parametrized by their number of nodes and the probability of adding an edge in the graph. Barabasi-Albert model is used to generate scale-free graphs and are defined by the number of nodes and an attachment parameter. Edges are added preferentially to nodes having a higher degree.
%
%\item[Graph coloring (DIMACS)] 
%
%\item[Maximum clique problem (DIMACS)] 
%
%
%\end{description}

\subsection{Experimental Protocol}

MISP  instances were generated using  the Barabasi-Albert (BA) model \cite{albert2002statistical}. Such a model
is commonly used for generating  real-world and scale-free graphs.  They are defined by the number of nodes ($n$) and an attachment parameter ($\nu$).   The greater is $\nu$, the denser is the graph.   Edges are added preferentially to nodes having a higher degree. 
Training has been carried out on Compute Canada Cluster\footnote{https://www.computecanada.ca/research-portal/}.
Training time is limited to 20 hours, memory consumption to 64 GB and one GPU (NVIDIA P100 Pascal, 12GB HBM2 memory) is used.
For each configuration, the training is done using 1000 generated random BA graphs (between 90 and 100 nodes) that are refreshed every 5000 iterations. Different models with a specific value for the attachment parameter ($\nu=\{2, 4, 8, 16 \}$) are trained. The model selected is the one giving the best average reward on a validation set composed of 100 graphs having the same configuration as the training graphs. The training time required to get this model is dependent on the configuration considered.
It varies between 20 minutes for the best case and 6 hours for the worst case.
At the first time, testing is carried out on 100 other random graphs of the same size and having the same attachment parameter as for the training. Other configurations are then considered. 
Performance profiles \cite{dolan2002benchmarking} are used for comparing the approaches. This tool provides a synthetic view on how an approach performs compared to the others tested.
The metric considered  is the optimality gap (i.e. the relative distance between the bound and the optimal solution). 
%Table \ref{par_table} gives a summary of the hyper-parameters that we used for the training.

%
%\begin{table}[!ht]
%\centering
%
%
%	\begin{adjustbox}{max width=\columnwidth}
%\begin{tabular}{|l|l|c|c|}
% \hline
%\multicolumn{2}{|c|}{Parameters} & \texttt{RL-UB}  & \texttt{RL-LB}   \\
%\hline
%Batch size & $B$  & &   \\
%Training graph size & $\underline{n},\overline{n}$  &  &    \\
%Learning rate & $\alpha$ &  &   \\
%Disc. factor & $\gamma$ &  &    \\
%\# Iterations & $K$  &  &    \\
%\# Training graph & $m$  &  &    \\
%Reward scaling & $\rho$ &  &   \\
%Epsilon-greedy  & $\epsilon$ & &    \\
%Store size (MB) & $|\mathcal{D}|$ & &    \\
%\hline
%\end{tabular}
%\end{adjustbox}
%\caption{Hyper-parameter values of the different models.}
%\label{par_table}
%\end{table}

%Let $A$ be the set of the approaches we are considering, $I$ the set of tested instances and $m_{i,a}$ the metric we want to access obtained when applying the approach $a \in A$ on the instance $i \in I$. We define the performance ratio of an approach $a$ on the instance $i$ as $r_{i,a}= \frac{m_{i,a}}{m^\star_i}$ where $m^\star_i$ denotes the optimal solution on instance $i$. The performance profile of an approach $a$ is  a cumulative distribution function of the performance of $s$ compared to the other approaches: $\rho_a(\tau) = \frac{1}{|I|} \times  \big|  \{ i \in I  ~ | ~ r_{i,a} \leq \tau \}  \big|$. 

%\begin{equation}
%\label{eq:pp}
%\rho_a(\tau) = \frac{1}{|I|} \times  \Big|  \big\{ i \in I  ~ \big| ~ r_{i,a} \leq \tau \big\}  \Big|
%\end{equation}
% 

Our model is implemented upon the code of Dai et al.\footnote{https://github.com/Hanjun-Dai (graph\_comb\_opt)} for the learning part and upon the code of Bergman et al. \cite{bergman2013optimization} for building the DDs of the MISP instances. Evaluation of the different orderings is also done using this software. The learning is done using Adam optimizer \cite{kingma2014adam}. Library  \texttt{networkX} \cite{hagberg2008exploring} is used for generating the random graphs. For the reproducibility of results, the implementation of our 
approach is available online\footnote{https://github.com/qcappart/learning-DD}. Optimal solutions of the MISP instances and the linear relaxations have been obtained using CPLEX 12.6.3.

% Link of graph_comb \url{github.com/Hanjun-Dai/graph_comb_opt}
\subsection{Results}

The goal of the experiments is to show the adequacy of our approach for computing both upper and lower bounds in different scenarios commonly considered in practice. 

%More specifically, the following analysis are carried out for both restricted and relaxed DDs:

%\begin{enumerate}
%\item The impact of the BDD width chosen for the training and the selection of the most appropriated one.
%\item The performance of the approach compared to the other heuristic for computing the bounds. Different distributions for the graphs are considered.
%\item The stability and the execution time of the approach when a larger width than the one used for the training is considered during the evaluation.
%\item The stability and the execution time of the approach when larger graphs than the one used for the training are considered in the test set.
%\item The robustness of the approach when the evaluation is done on graphs following a different distribution than during the training.
%\end{enumerate}

\begin{figure*}[!ht]
 \centering
    \subfloat[Testing on $w=2$.]{
      \includegraphics[width=0.24\textwidth]{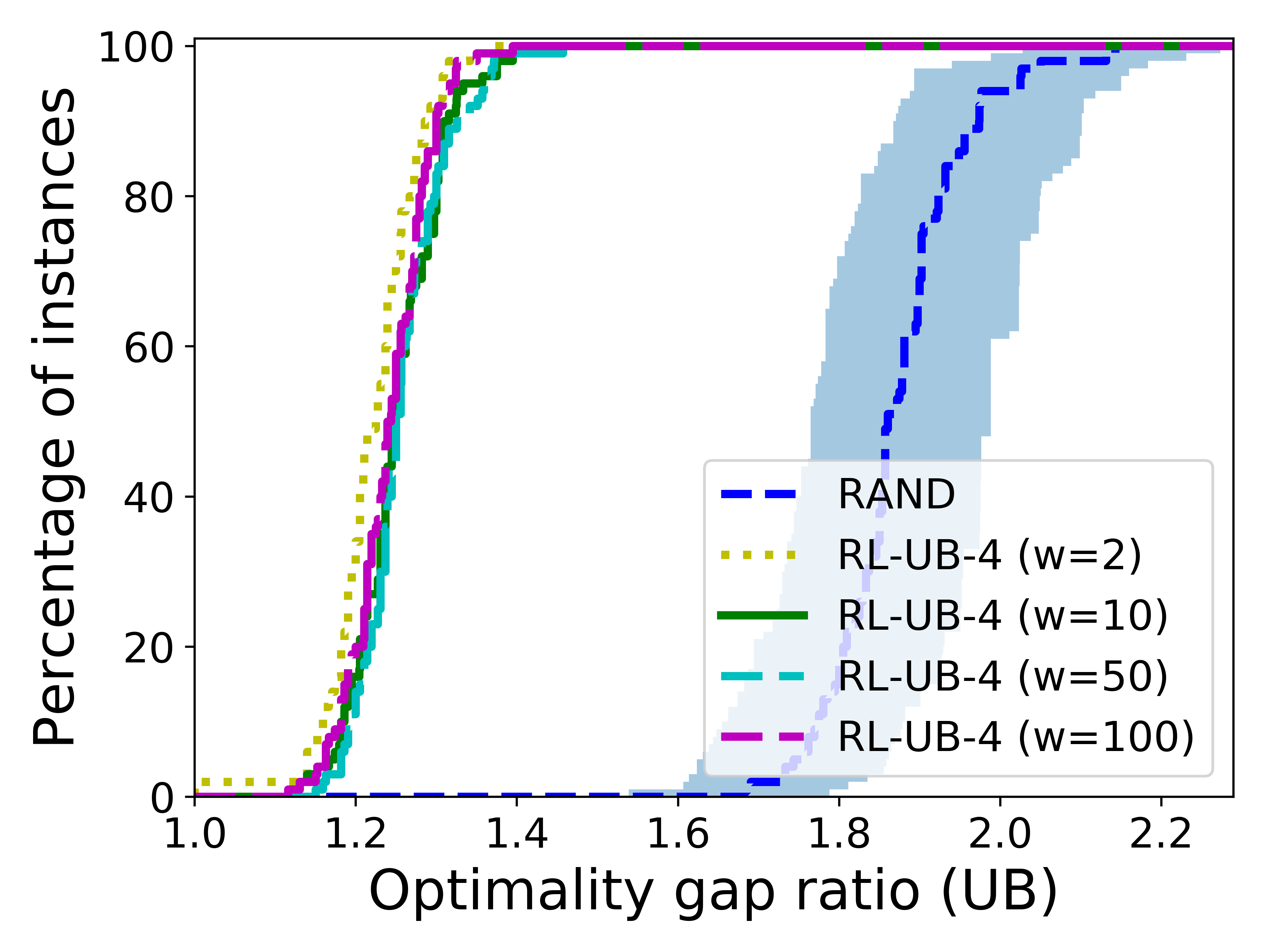}
      \label{sub:tw1}
                         }
    \subfloat[Testing on $w=10$.]{
      \includegraphics[width=0.24\textwidth]{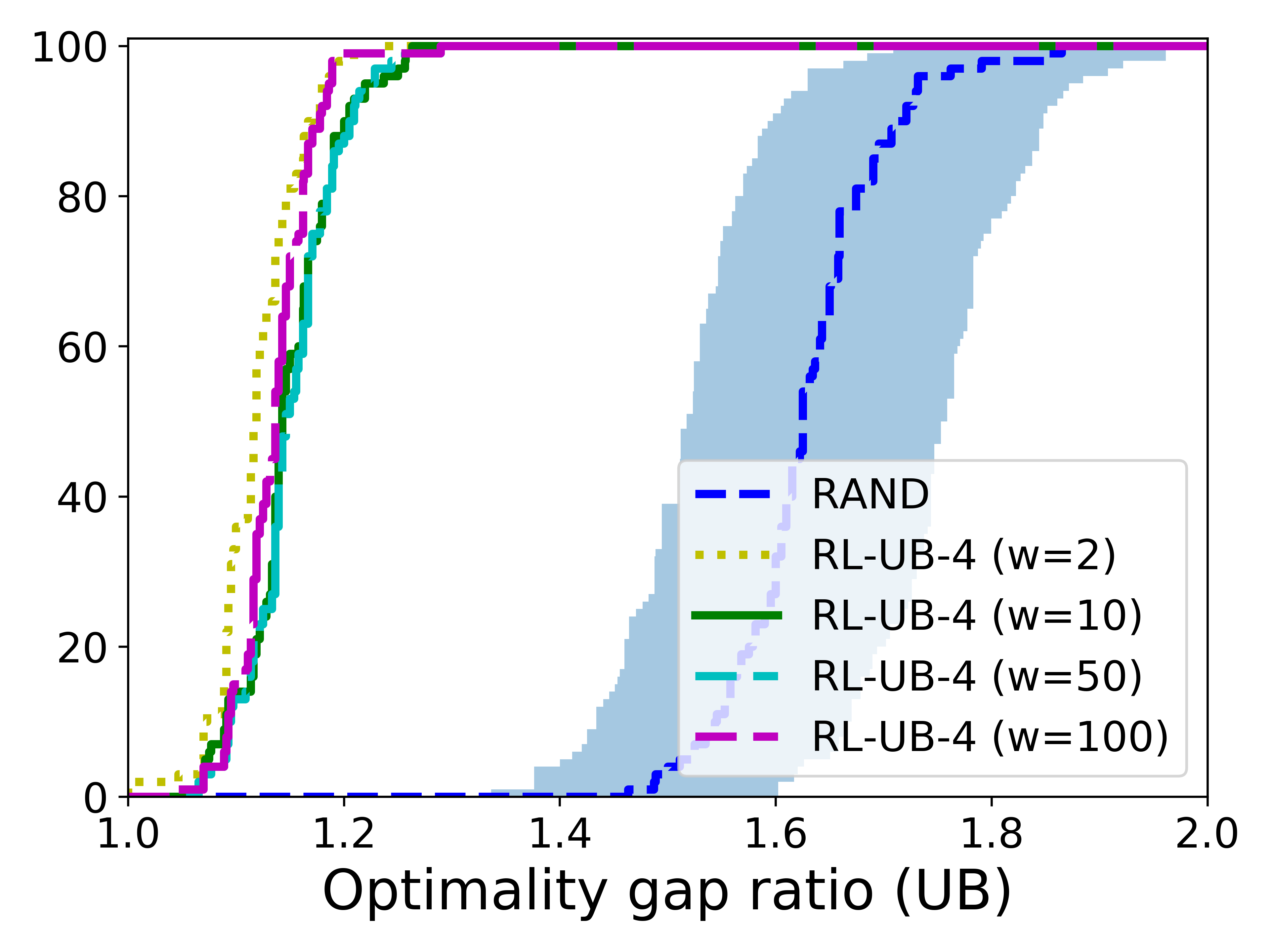}
      \label{sub:tw2}
                         } 
         \subfloat[Testing on $w=50$.]{
      \includegraphics[width=0.24\textwidth]{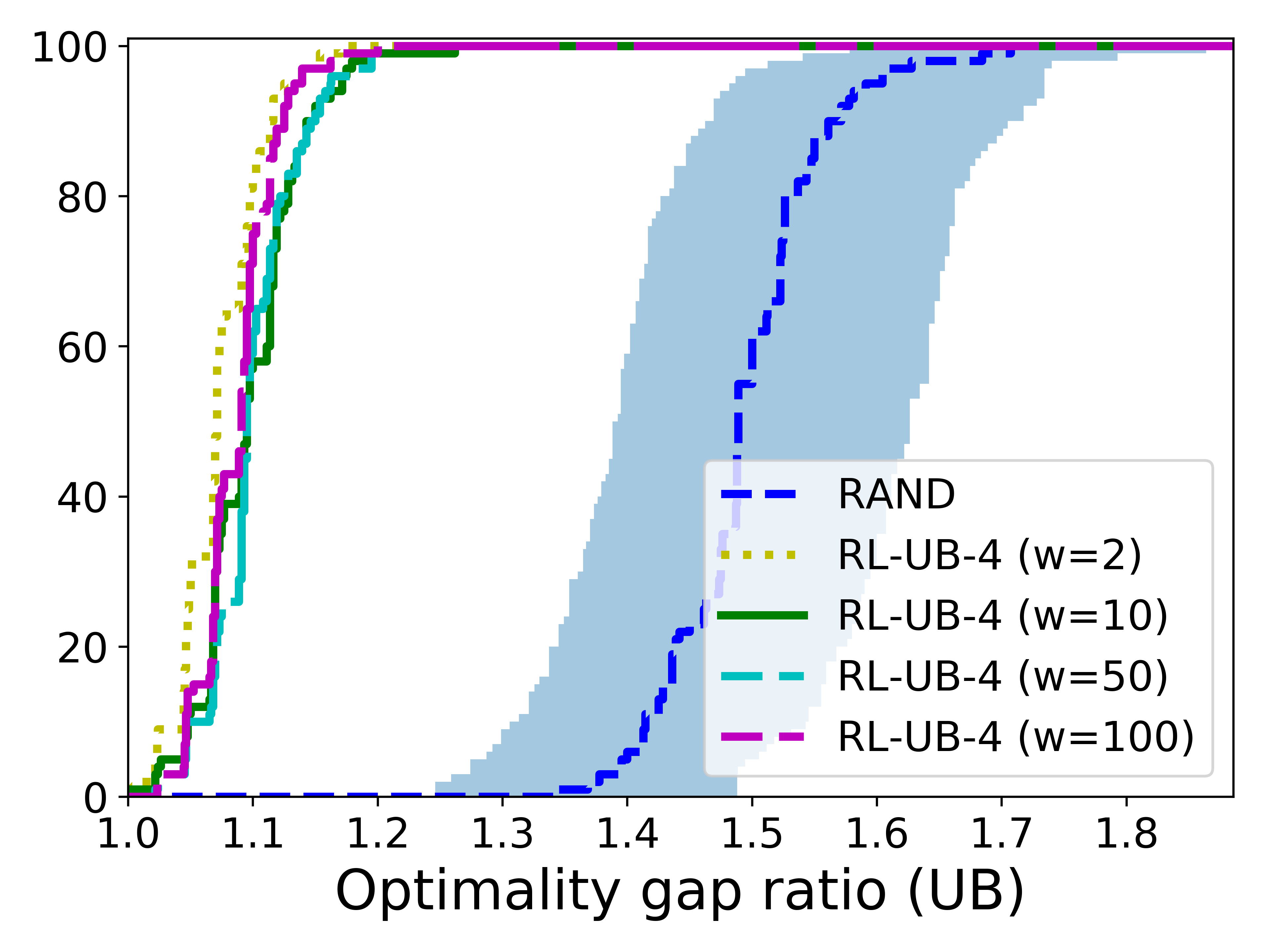}
      \label{sub:tw3}
                         }
         \subfloat[Testing on $w=100$.]{
      \includegraphics[width=0.24\textwidth]{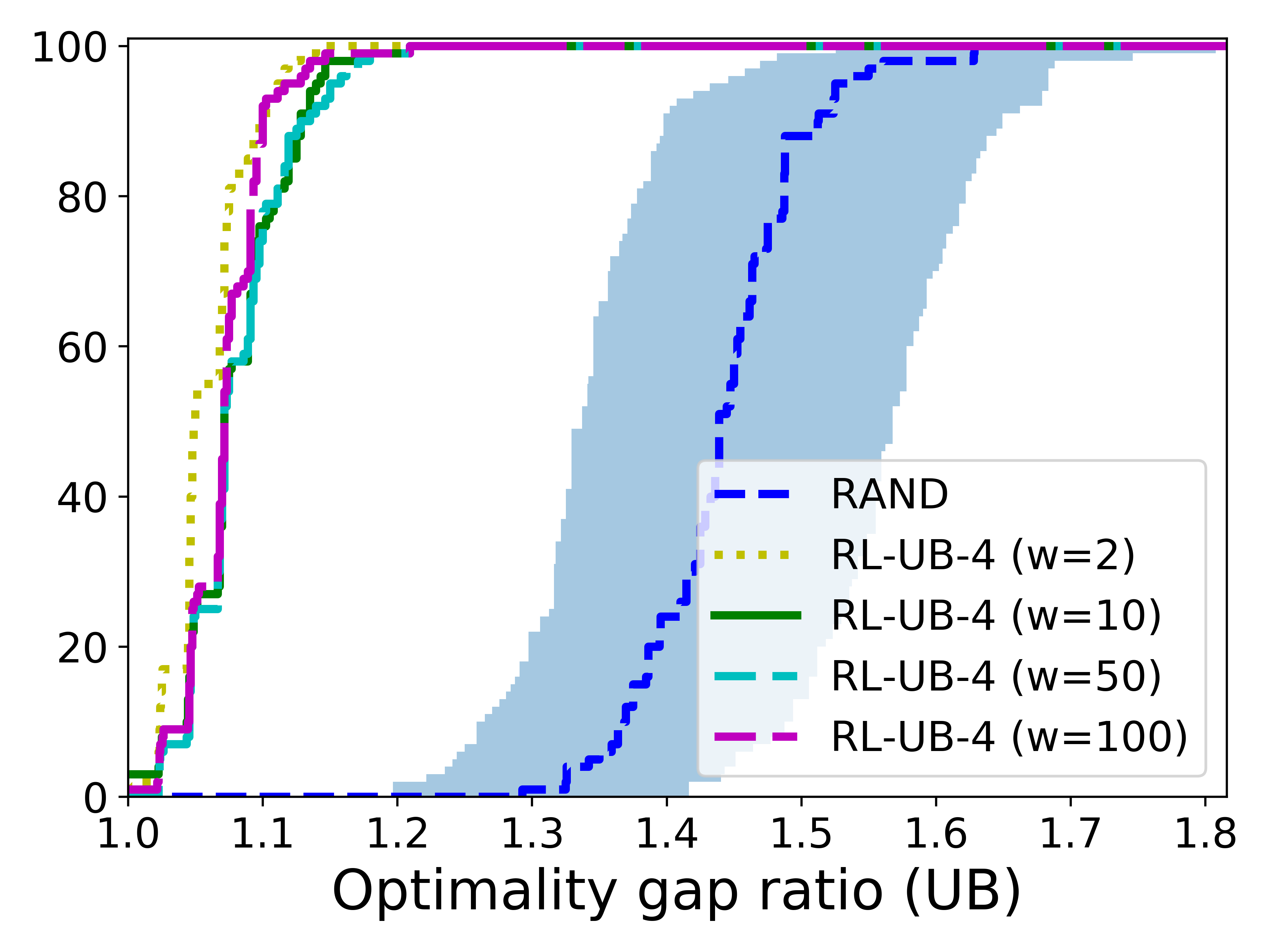}
      \label{sub:tw4}
                         }                        
    \caption{Performance profiles of model trained with different widths for relaxed DDs.}
    \label{fig:training-width}
\end{figure*}

\begin{figure*}[!ht]
 \centering
\subfloat[$\nu=2$.]{
      \includegraphics[width=0.24\textwidth]{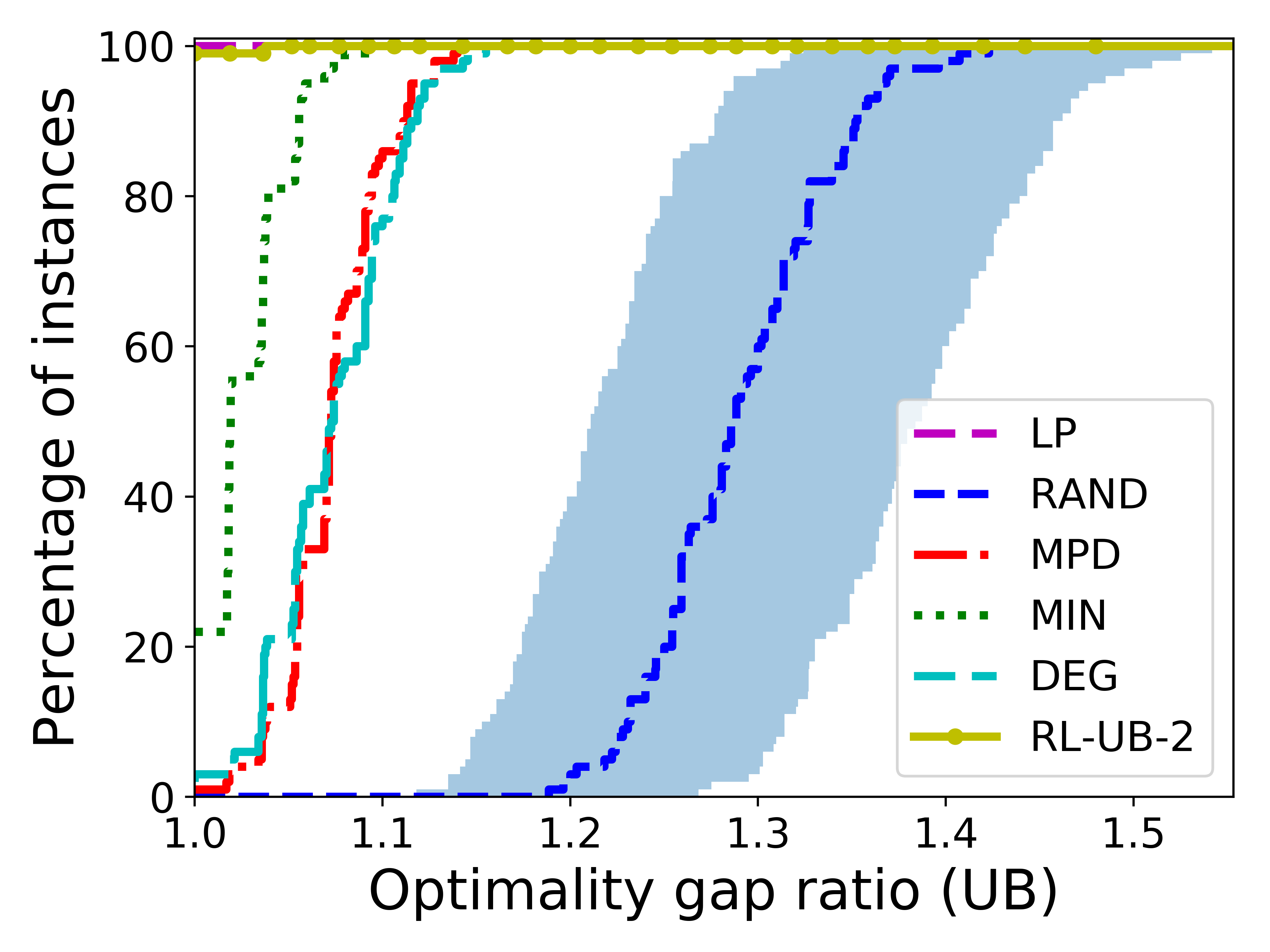}
      \label{sub:prel1}
                         }
    \subfloat[$\nu=4$.]{
      \includegraphics[width=0.24\textwidth]{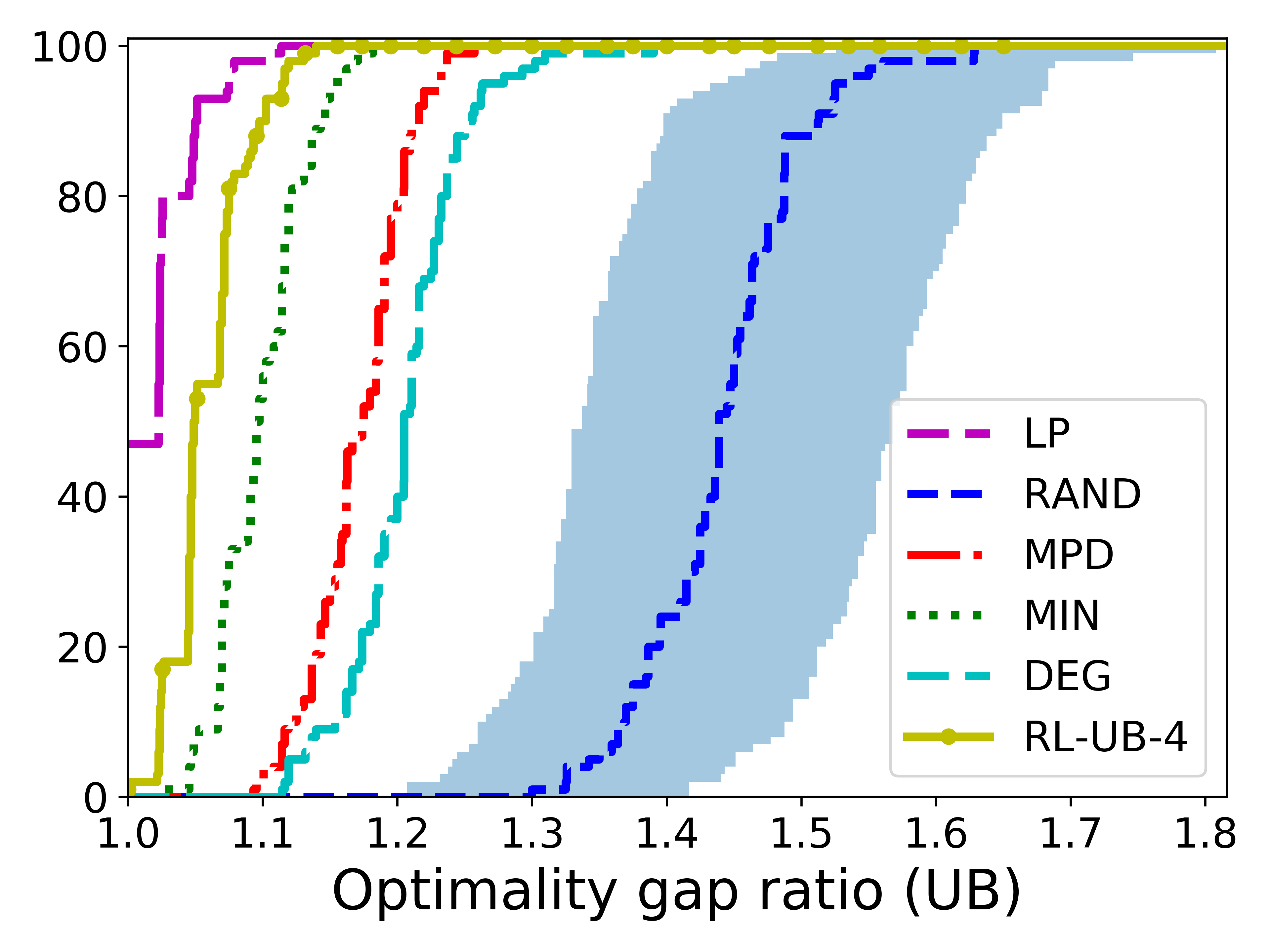}
      \label{sub:prel2}
                         } 
         \subfloat[$\nu=8$.]{
      \includegraphics[width=0.24\textwidth]{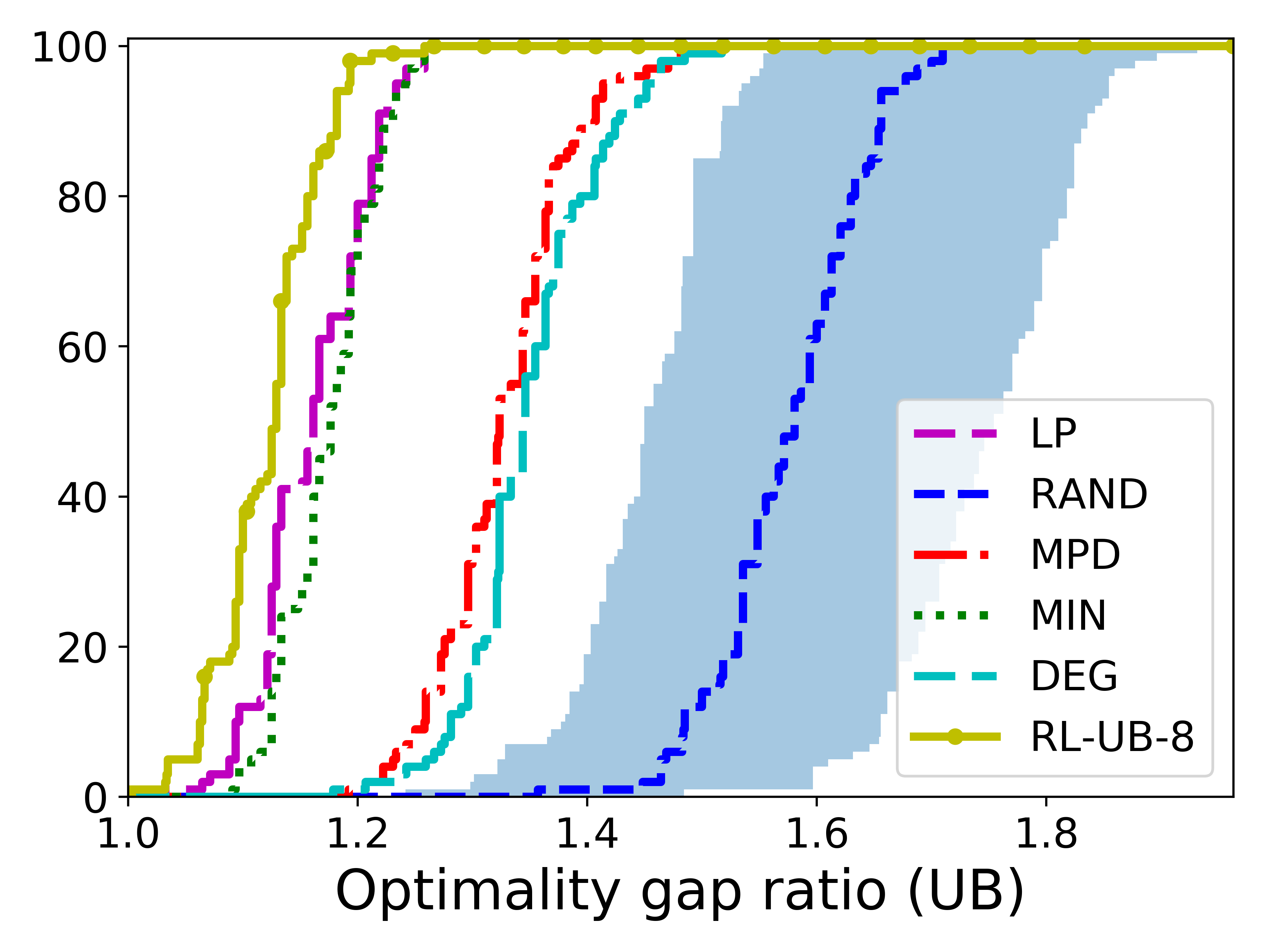}
      \label{sub:prel3}
                         }
         \subfloat[$\nu=16$.]{
      \includegraphics[width=0.24\textwidth]{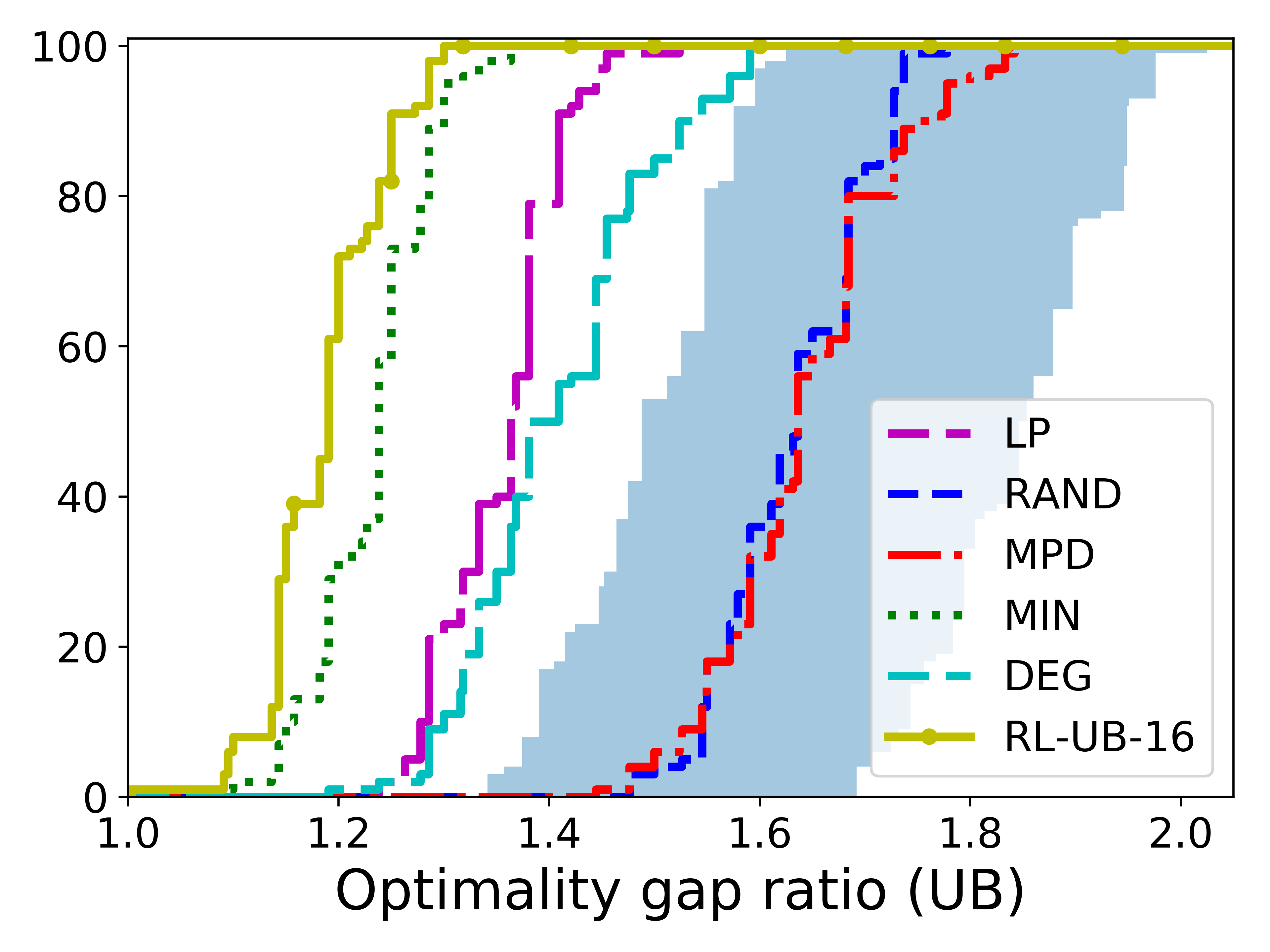}
      \label{sub:prel4}
                         } 
    \caption{Performance profiles on graphs of different distributions ($\nu$) for relaxed  DDs ($w = 100$).}
    \label{fig:perf-relaxed}
    
\end{figure*}

\begin{figure*}[!ht]
 \centering
    \subfloat[$\nu=2$.]{
      \includegraphics[width=0.24\textwidth]{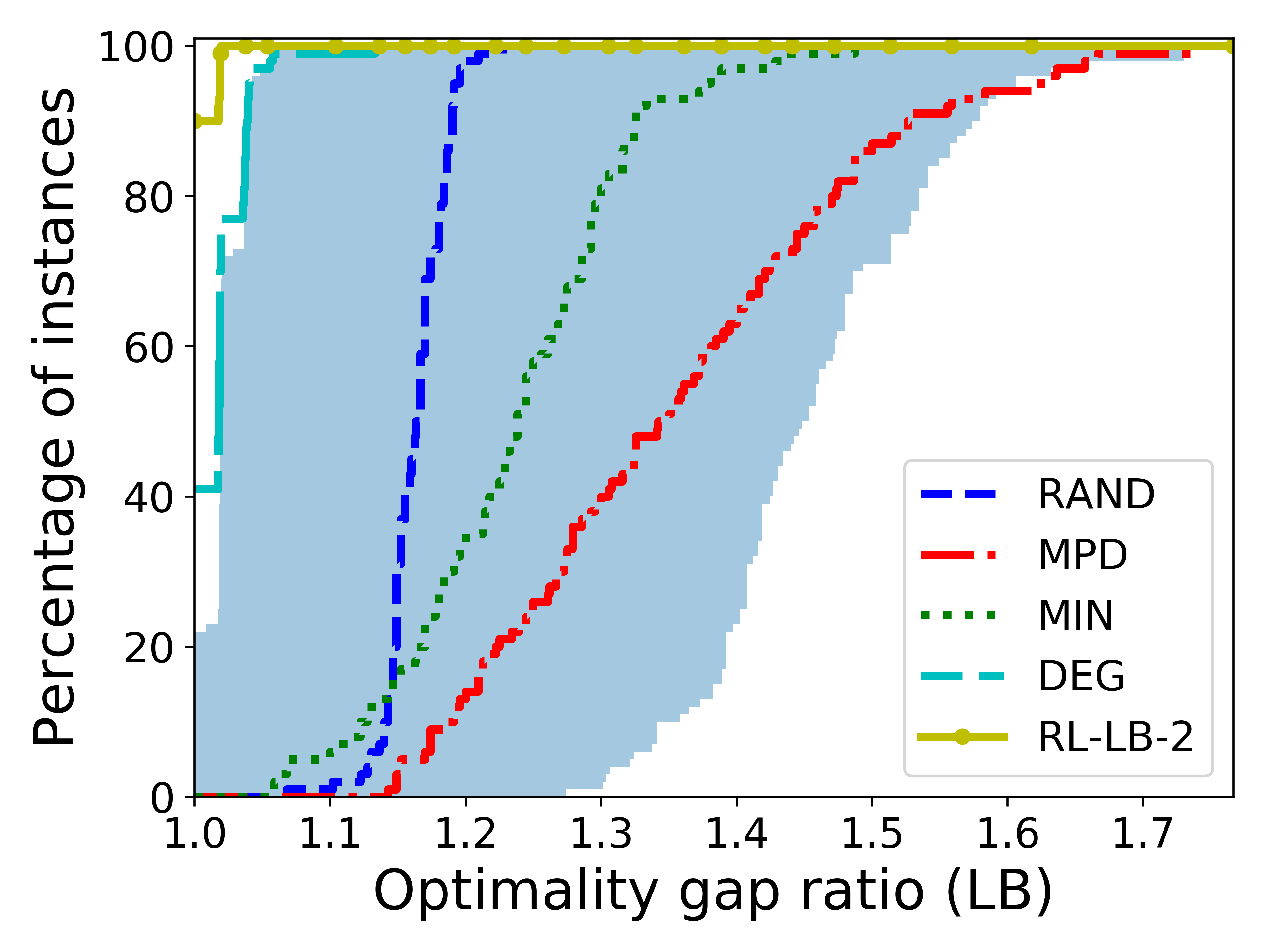}
      \label{sub:pres1}
                         }
    \subfloat[$\nu=4$.]{
      \includegraphics[width=0.24\textwidth]{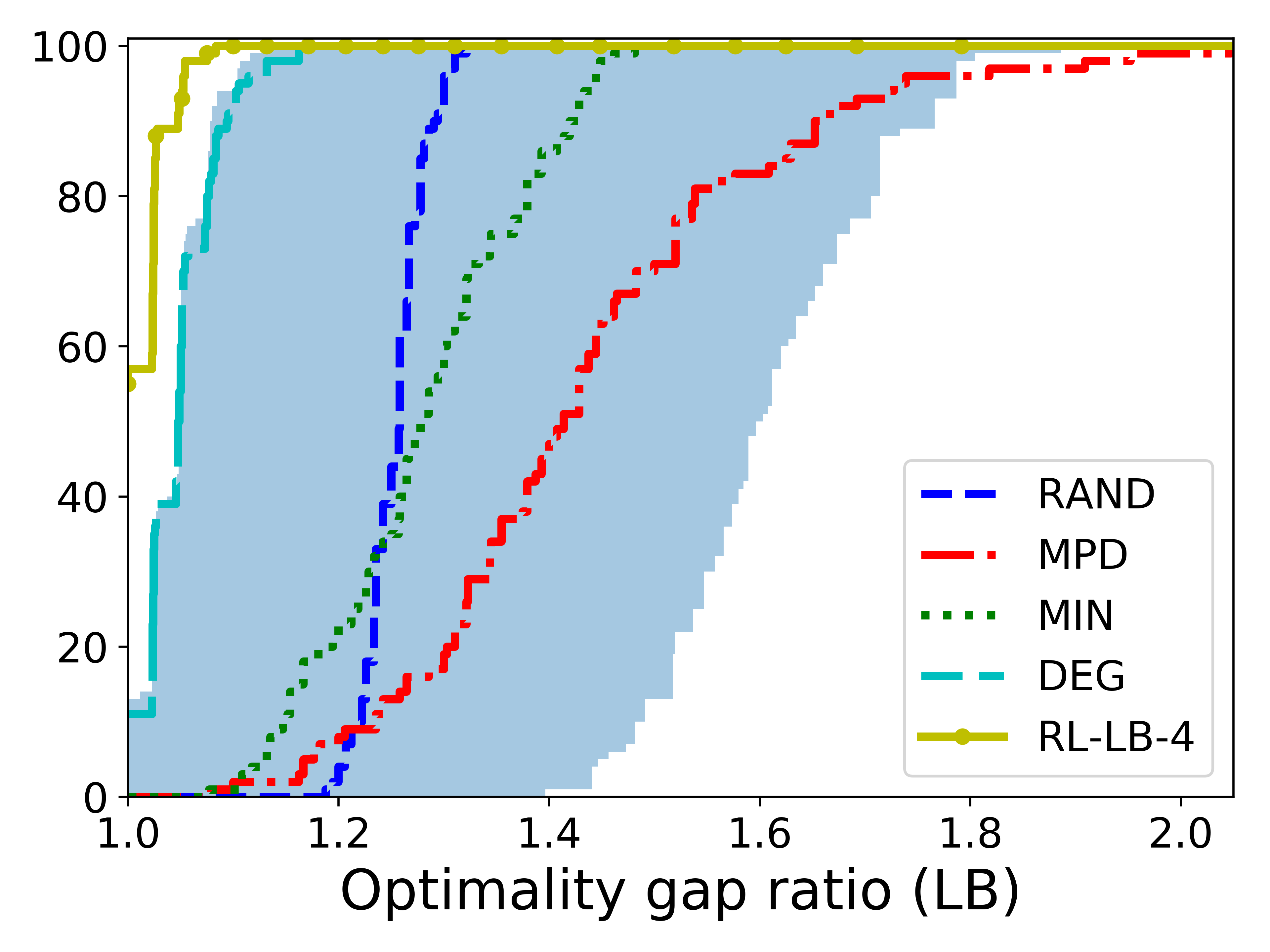}
      \label{sub:pres2}
                         } 
         \subfloat[$\nu=8$.]{
      \includegraphics[width=0.24\textwidth]{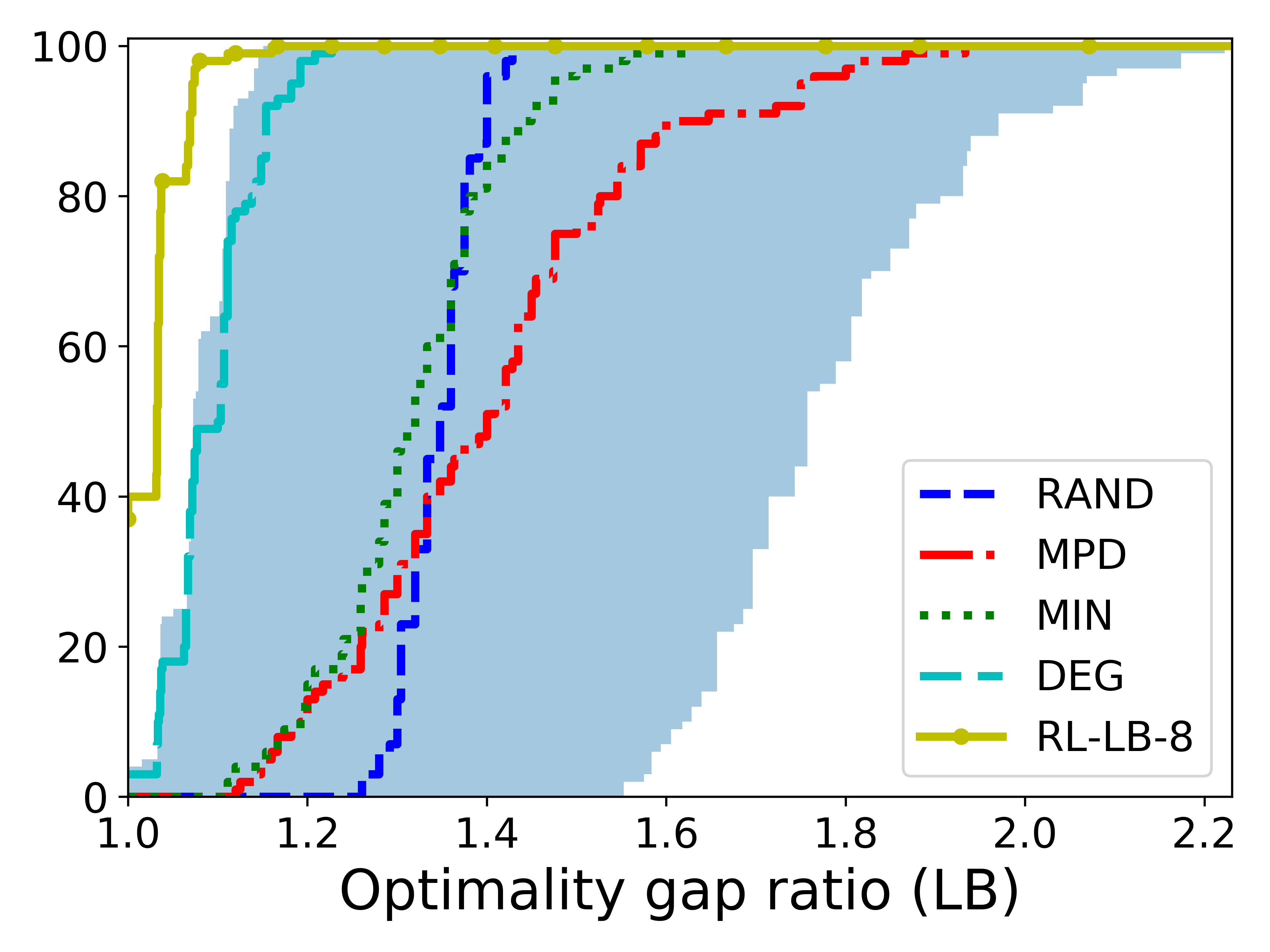}
      \label{sub:pres3}
                         }
         \subfloat[$\nu=16$.]{
      \includegraphics[width=0.24\textwidth]{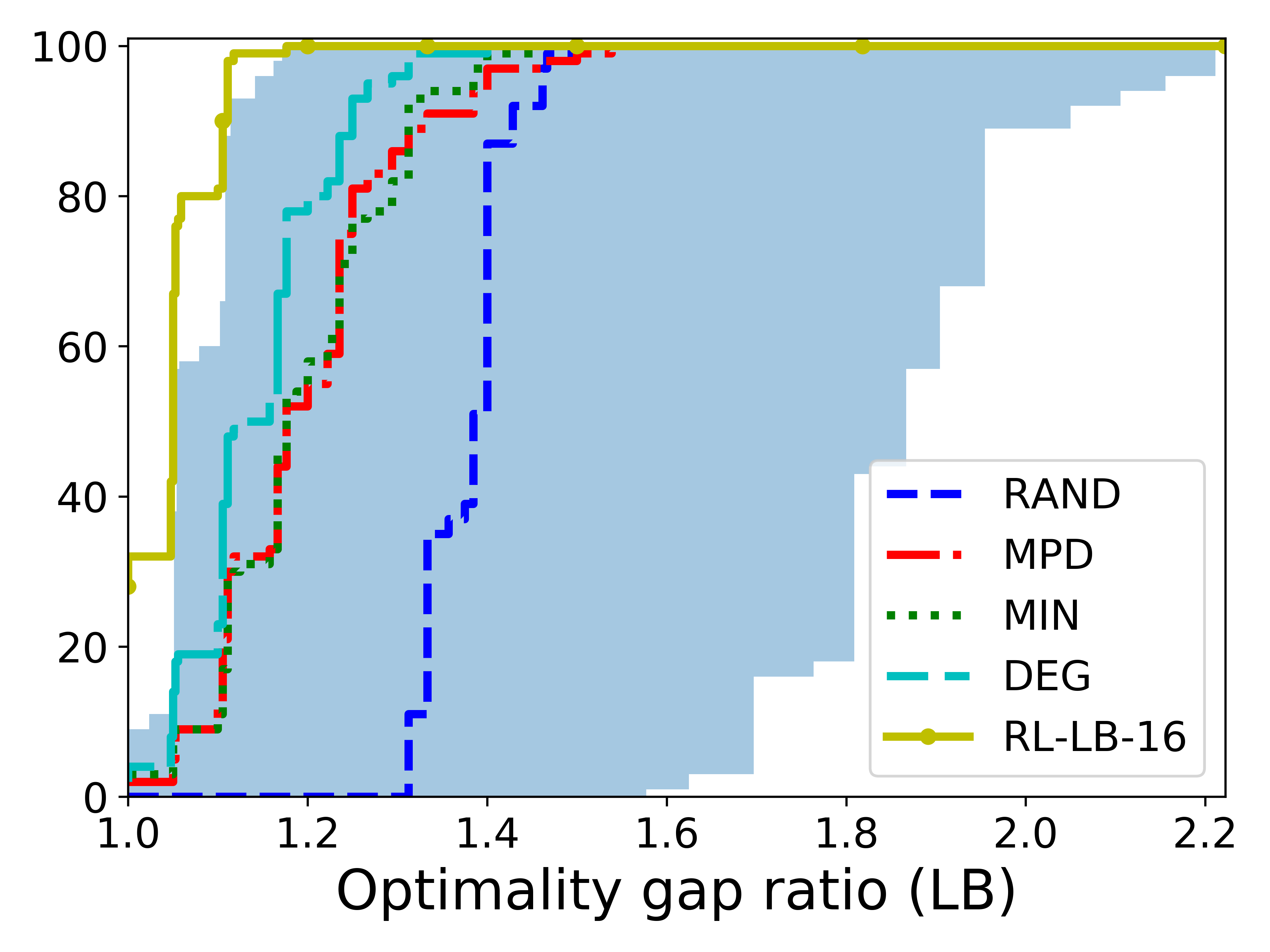}
      \label{sub:pres4}
                         }                        
    \caption{Performance profiles on graphs of different distributions ($\nu$) for restricted  DDs ($w = 2$).}
    \label{fig:perf-restricted}
    
\end{figure*}

\subsubsection{Evaluating the DD Width for Training}

The first set of experiments aim to determine the best DD maximal width ($w$) for training the model. Let us first consider $\nu=4$ for the attachment parameter as in \cite{khalil2017learning}. We trained four models ($w=\{2,10,50,100\}$) for relaxed DDs (\texttt{RL-UB-4}), and tested the models using the same values of $w$.  Figure \ref{fig:training-width} shows the performance profiles of the models when evaluated on relaxed DDs of a various width. Random ordering (\texttt{RAND}) is also reported and is outperformed by the four  models. The shaded area represent the range of the \texttt{RAND} performance when considering the best and the worst solution obtained among the 100 trials.
Interestingly, these results suggest that the width chosen for the training has a negligible impact on the quality of the model, even when the width considered during the testing is different than that for the training. 
As computing small-width DDs is less computationally expensive than those with larger widths, we select the model trained with a width of 2 for the remainder of the experiments on MISP. Concerning restricted DDs, as shown in the next set of experiments (Figures \ref{sub:pres1}-\ref{sub:pres4}), lower bounds close to the optimal solutions are already obtained with small-width DDs ($w=2$). This independence of the width chosen during the training is the most surprising result that we get through our set of experiments. 
Perhaps one reason of this stability is due to the merging heuristic that remains the same in all the configuration, but an in-depth explanation of these results is still an open question.

%\begin{figure}[h]
%  \begin{center}
%    \subfloat[BA with $m=2$.]{
%      \includegraphics[width=0.5\columnwidth]{performance-profile}
%      \label{sub:renonc}
%                         }
%    \subfloat[BA with $m=4$.]{
%      \includegraphics[width=0.5\columnwidth]{performance-profile/pp-ba-2-4.png}
%      \label{sub:popul}
%                         } \\
%         \subfloat[BA with $m=8$.]{
%      \includegraphics[width=0.5\columnwidth]{performance-profile}
%      \label{sub:popul}
%                         }
%         \subfloat[BA with $m=16$.]{
%      \includegraphics[width=0.5\columnwidth]{performance-profile}
%      \label{sub:popul}
%                         } \\
%    \subfloat[ER with $p=0.1$.]{
%      \includegraphics[width=0.5\columnwidth]{performance-profile}
%      \label{sub:renonc}
%                         }
%    \subfloat[BA with $p=0.2$.]{
%      \includegraphics[width=0.5\columnwidth]{performance-profile}
%      \label{sub:popul}
%                         } \\
%         \subfloat[BA with $p=0.3$.]{
%      \includegraphics[width=0.5\columnwidth]{performance-profile}
%      \label{sub:popul}
%                         }
%         \subfloat[BA with $p=0.4$.]{
%      \includegraphics[width=0.5\columnwidth]{performance-profile}
%      \label{sub:popul}
%                         }                         
%    \caption{Performance profiles of different configurations for relaxed DDs having a with of 2.}
%    \label{fig:renonculacees}
%  \end{center}
%\end{figure}

%

\subsubsection{Comparison with Other Methods}

Our approach is now compared to the other variable ordering heuristics using BA graphs having a varied density ($\nu=\{2, 4, 8, 16 \}$). 
A specific model is trained for each distribution for both relaxed (\texttt{RL-UB-}$\nu$) and restricted DDs  (\texttt{RL-LB-}$\nu$).
Evaluation is done on graphs following the same distribution as those used in training.
Results are presented in Figures \ref{sub:prel1}-\ref{sub:prel4} for relaxed DDs having a width of 100. 
In all the configurations tested, our approach provides a better upper bound than the \texttt{RAND}, \texttt{MIN}, \texttt{MPD}  and \texttt{DEG} heuristics.
For sparsest graphs (Figure \ref{sub:prel1}), the optimal solution is reached for almost all the instances. When the graphs are relatively sparse (Figure \ref{sub:prel2}), the linear relaxation provides the best bound. 
 However, this trend decreases as the density of the graphs grows (Figures \ref{sub:prel3} and \ref{sub:prel4}). For these graphs, our model gives the best performance  for all the instances.
Results for restricted DDs with a width of 2 are depicted on Figures \ref{sub:pres1}-\ref{sub:pres4}. Again, our model has the best results over those tested and provides stronger lower bounds,  close to the optimal solution. Optimality is reached for $\approx 90 \%$ of the easiest instances ($\nu=2$) and for $\approx 30 \%$ of the hardest ones ($\nu=16$).

\subsubsection{Analysis of Width Evolution}

\begin{figure}[ht]
 \centering
    \subfloat[Evolution of width.]{
      \includegraphics[width=0.48\columnwidth]{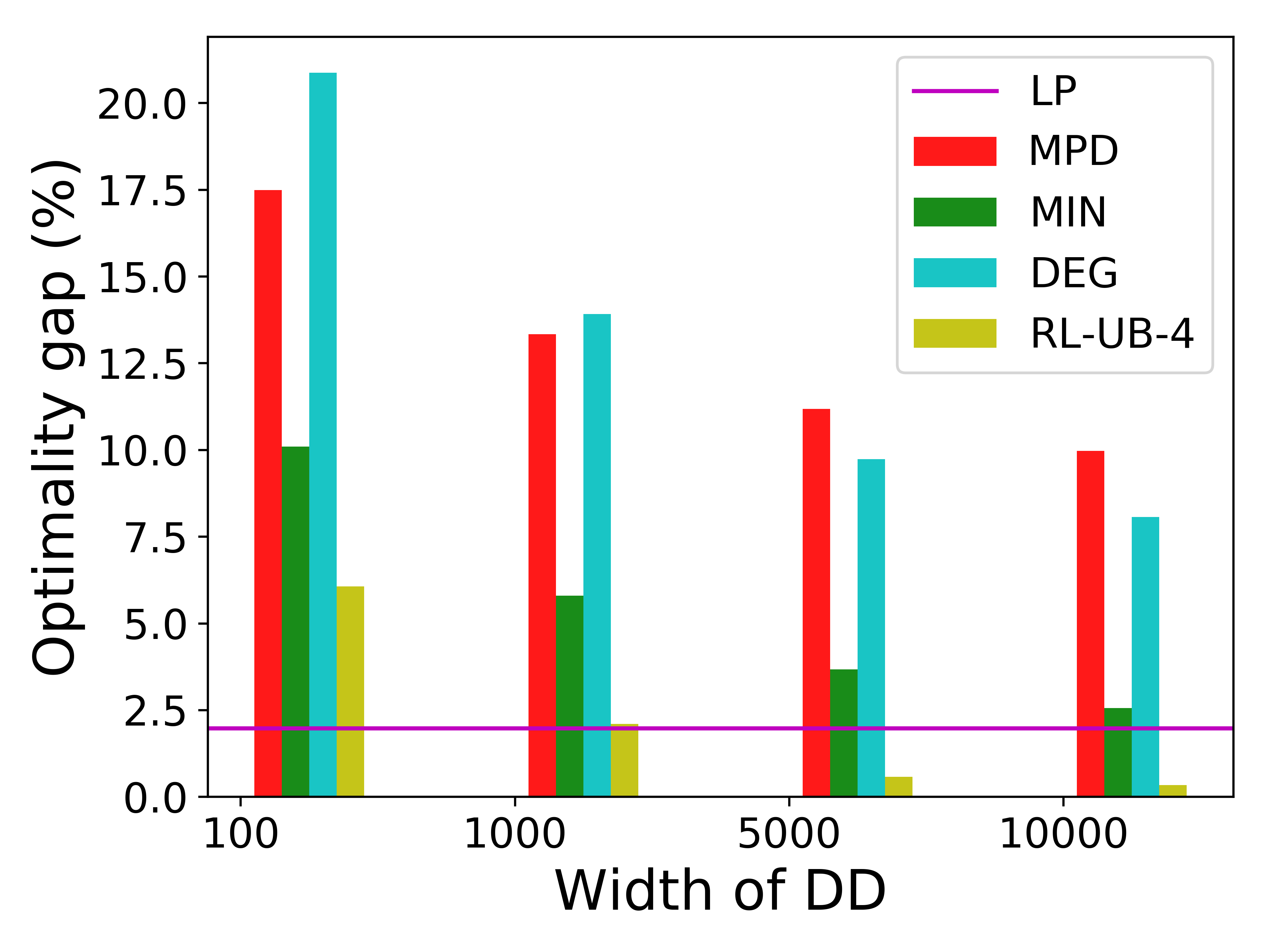}
      \label{sub:we1}
                         }
    \subfloat[Evolution of time.]{
      \includegraphics[width=0.48\columnwidth]{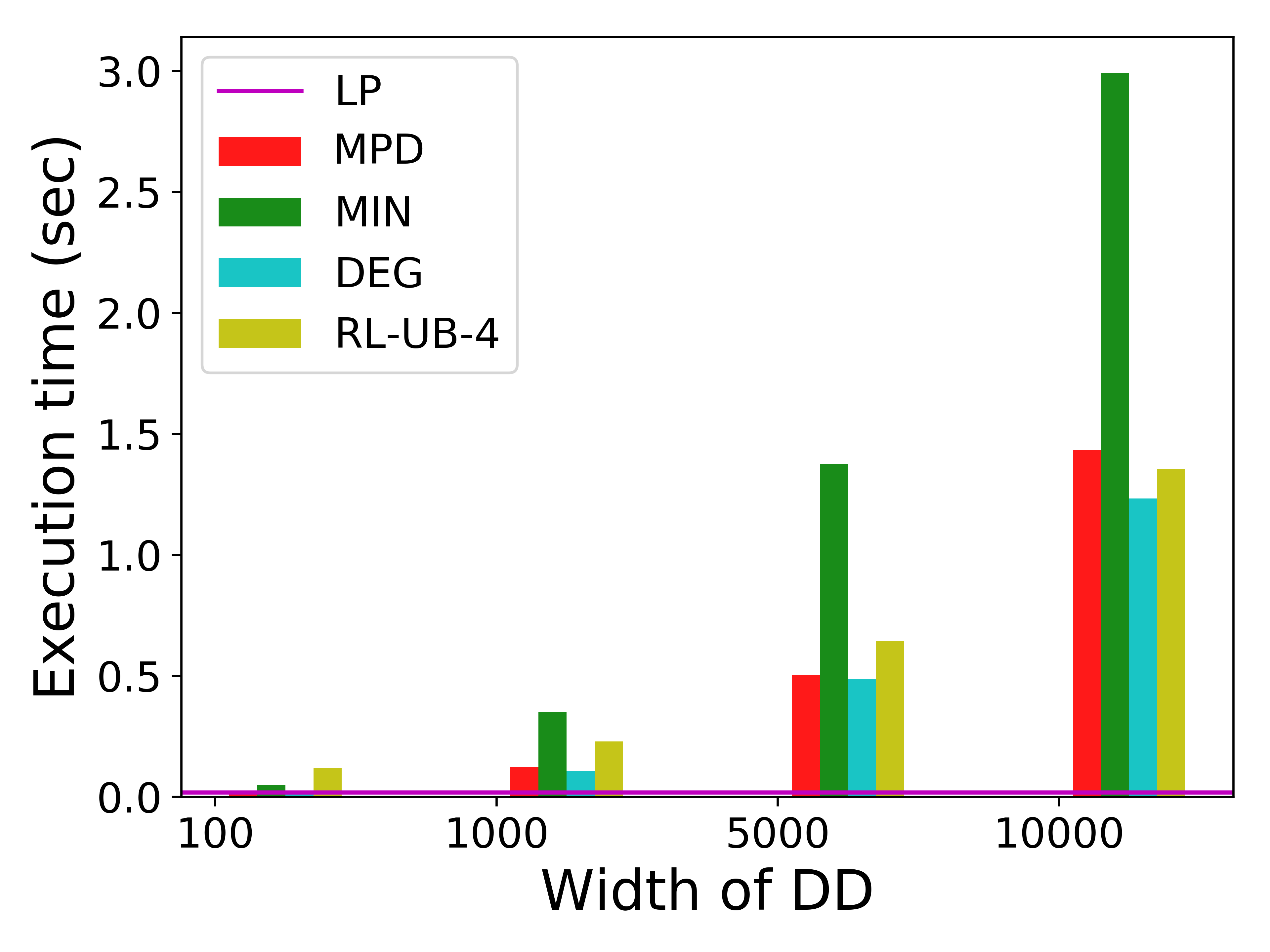}
      \label{sub:we2}
                         }                        
     \caption{Relaxed DDs of larger widths ($\nu=4$).}
    \label{fig:width-evol}
\end{figure}

Let us now consider the situation depicted in Figure \ref{sub:prel2} where \texttt{RL-UB-4} provides a worse bound than the linear relaxation of the problem. Figure \ref{sub:we1} depicts the evolution of the optimality gap when the model is tested on relaxed DDs of an increasingly larger width. As \texttt{RAND} provided results far outside the range of the other methods for relaxed DDs, we do not include it in the subsequent plots. 
The plot depicts that \texttt{RL-UB-4} remains better than the other ordering heuristics tested, and when the DD width is sufficiently large ($w>1000$) the LP relaxation bound is beaten and the optimal solution is almost reached ($w=10000$). 
Figure \ref{sub:we2} reports the execution time of the different methods. Concerning the RL model, only the time required for building the DD is reported. We do not report the training time on this experiment because it has to be amortized on all the graphs of the test set, which is dependent of the situation. The linear relaxation is the fastest method and is almost instantaneous. Concerning the orderings, \texttt{RL-UB-4}, \texttt{MPD} and \texttt{DEG} are static, and execution time for each generally increases similarly with the width, while \texttt{MIN} requires dynamically processing the nodes in a layer for determining the next vertex to insert. 
%This operations can be implement with time complexity $O(w\times |V|)$ in the worst case \textcolor{red}{do we need a reference? what about the time to learn the model?}.

%
%\begin{figure}[ht]
%  \begin{center}
%    \subfloat[Evolution of width.]{
%      \includegraphics[width=0.49\columnwidth]{results/width-evolution}
%      \label{sub:we1}
%                         }
%    \subfloat[Evolution of time.]{
%      \includegraphics[width=0.49\columnwidth]{results/time-evolution}
%      \label{sub:we2}
%                         }                        
%     \caption{Relaxed DDs of larger widths ($\nu=4$).}
%    \label{fig:width-evol}
%       \end{center}
%\end{figure}

\subsubsection{Analysis of Graph Size Evolution}

\begin{figure}[!ht]
   \centering
    \subfloat[Relaxed DDs ($w = 100$).]{
      \includegraphics[width=0.48\columnwidth]{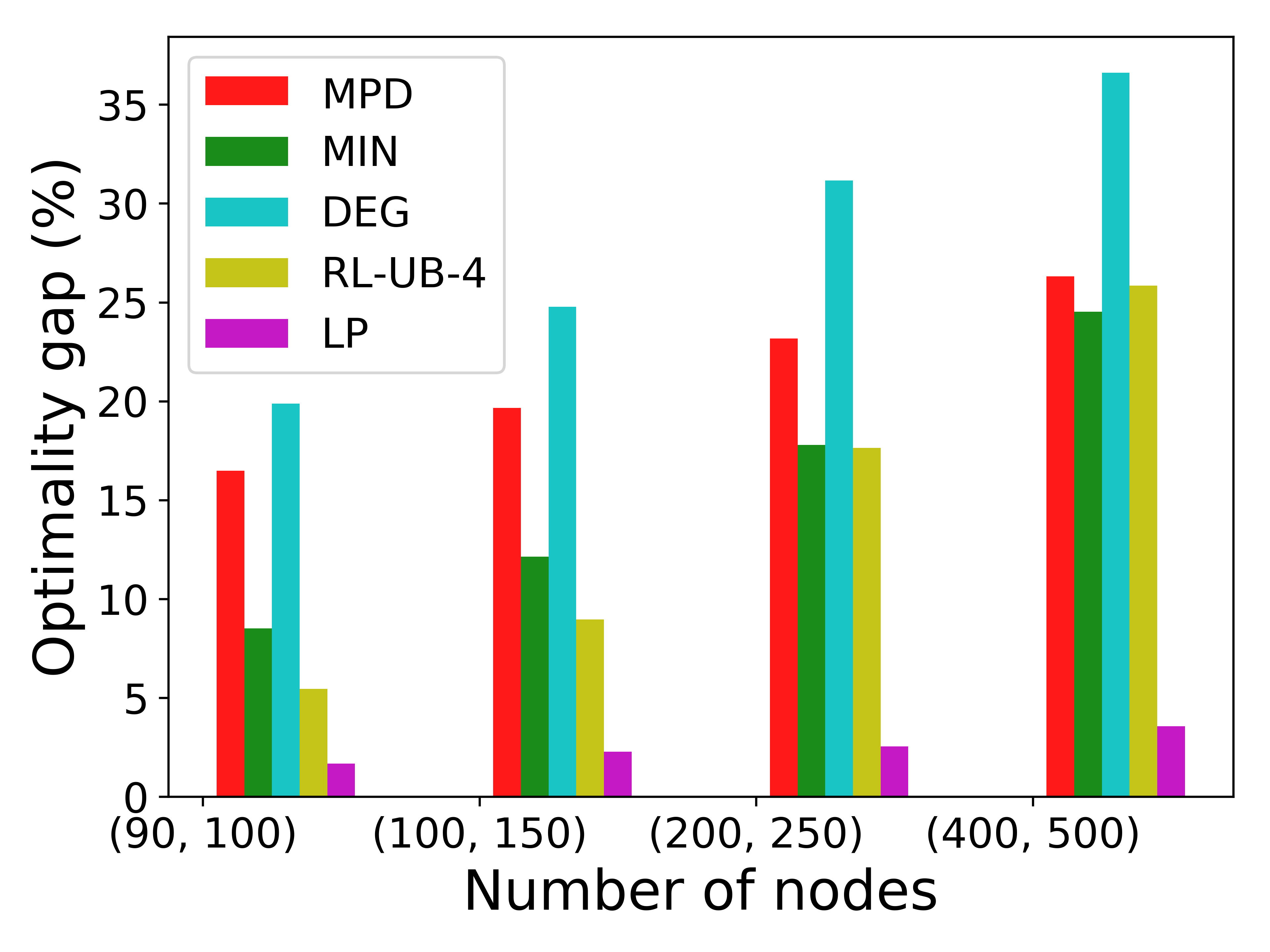}
      \label{sub:enrel1}
                         }
    \subfloat[Restricted DDs ($w = 2$).]{
      \includegraphics[width=0.48\columnwidth]{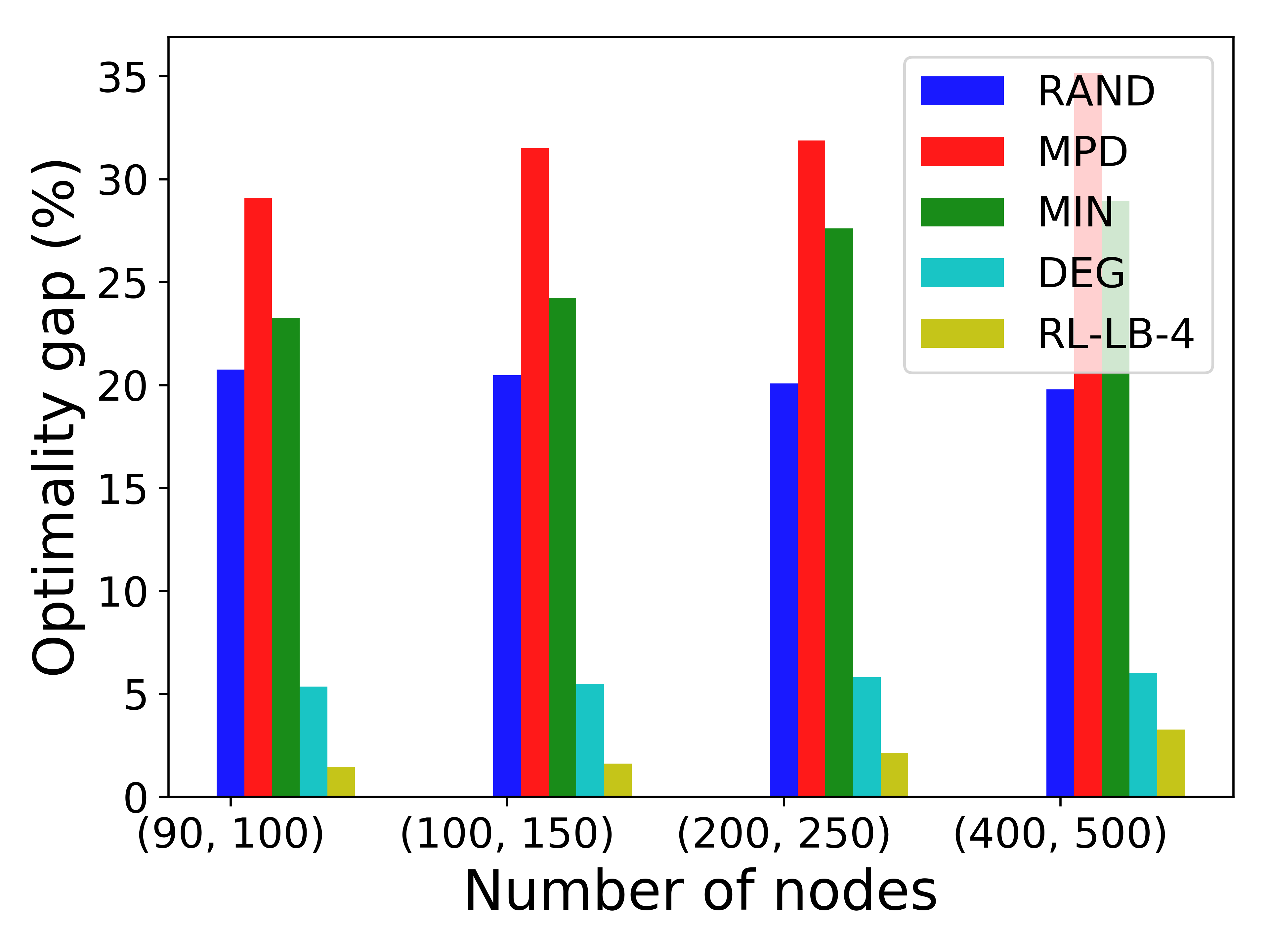}
      \label{sub:enrel2}
                         }                          
     \caption{Relaxed/Restricted DDs for larger graphs ($\nu=4$).}
    \label{fig:node-rel-evol}
\end{figure}

In a similar way, this set of experiments aim to analyze how the learned models perform when larger graphs are considered.
 Results in Figure \ref{sub:enrel1}  depict the optimality gap of the different approaches for relaxed DDs ($w=100$). 
 
 We can observe that the learned model remains robust against increases of the graph size although the  gap between the other orderings progressively decreases. 
When the graph size is far beyond the size used for the training, the model strives to generalize which indicates that training on larger graphs should be required.
The LP bounds for large graphs are out of range of DDs of this limited width.
The same experiment is carried out for restricted DDs and reported in Figure \ref{sub:enrel2}. Given that the optimality is reached even with small-width DDs, only  a width of 2 is considered. Here, \texttt{RL-LB-4} provides the best lower bound even for the largest graphs tested. This is consistent with other heuristics implemented through RL.

%\begin{figure}[ht]
%  \begin{center}
%    \subfloat[Relaxed DDs ($w = 100$).]{
%      \includegraphics[width=0.49\columnwidth]{results/nodes-evolution-relaxed-100}
%      \label{sub:enrel1}
%                         }
%    \subfloat[Restricted DDs ($w = 2$).]{
%      \includegraphics[width=0.49\columnwidth]{results/nodes-evolution-restricted-2}
%      \label{sub:enrel2}
%                         }                          
%     \caption{Relaxed/Restricted DDs for larger graphs ($\nu=4$).}
%    \label{fig:node-rel-evol}
%       \end{center}
%\end{figure}

%
%\begin{figure}[ht]
%  \begin{center}
%    \subfloat[Evolution of width.]{
%      \includegraphics[width=0.5\columnwidth]{results/nodes-evolution-restricted-2}
%      \label{sub:enres1}
%                         }
%    \subfloat[Evolution of time.]{
%      \includegraphics[width=0.5\columnwidth]{results/time-evolution}
%      \label{sub:enres2}
%                         } 
%                         
%     \caption{Analysis of the evolution of the number of nodes for BA graphs ($\nu=4$).}
%    \label{fig:node-res-evol}
%       \end{center}
%\end{figure}

\subsubsection{Performance on other Distributions}

\begin{figure}[!ht]
 \centering
  
    \subfloat[Relaxed DDs ($w = 100$).]{
      \includegraphics[width=0.48\columnwidth]{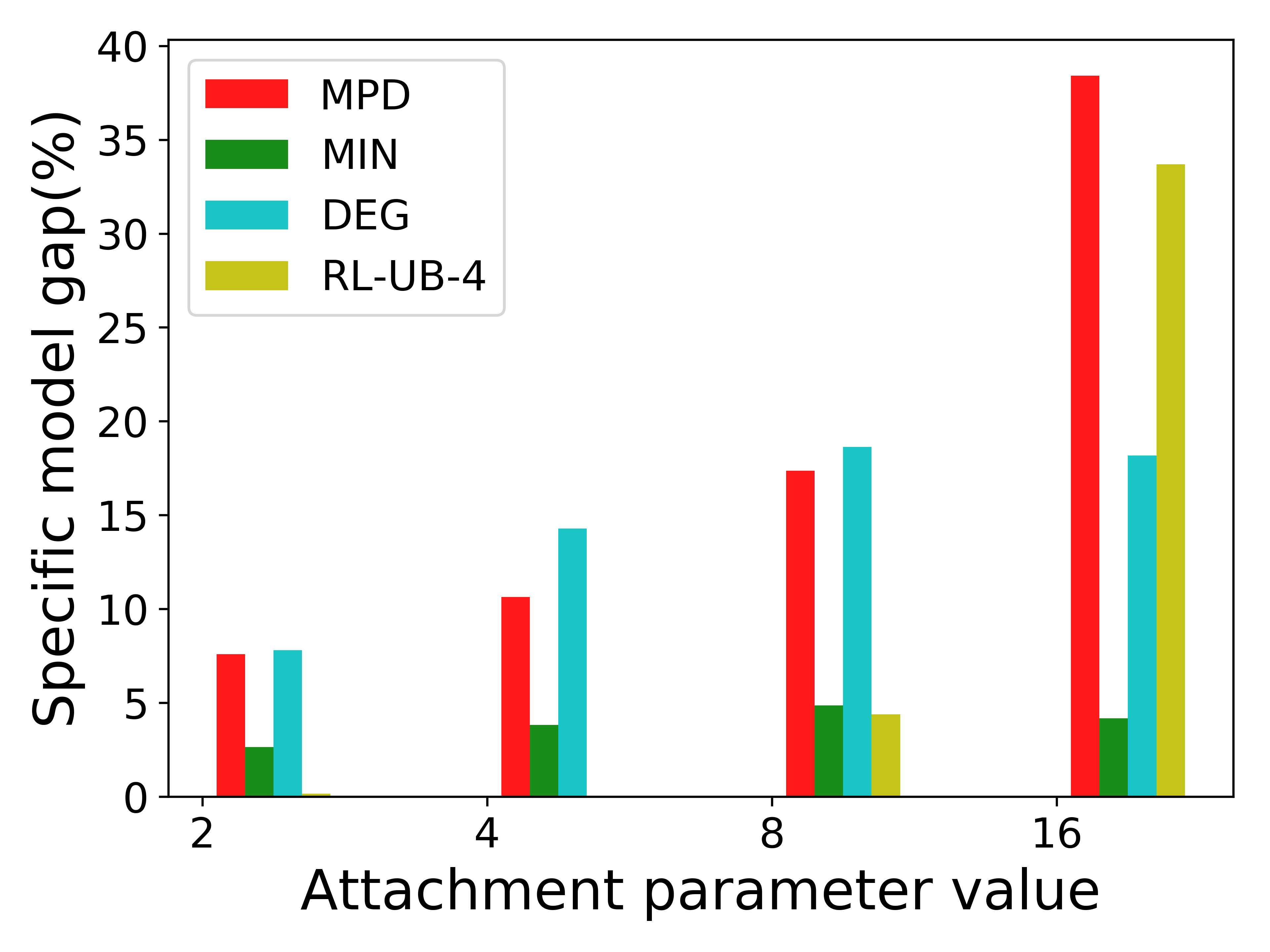}
      \label{sub:enres1}
                         }
    \subfloat[Restricted DDs ($w = 2$).]{
      \includegraphics[width=0.48\columnwidth]{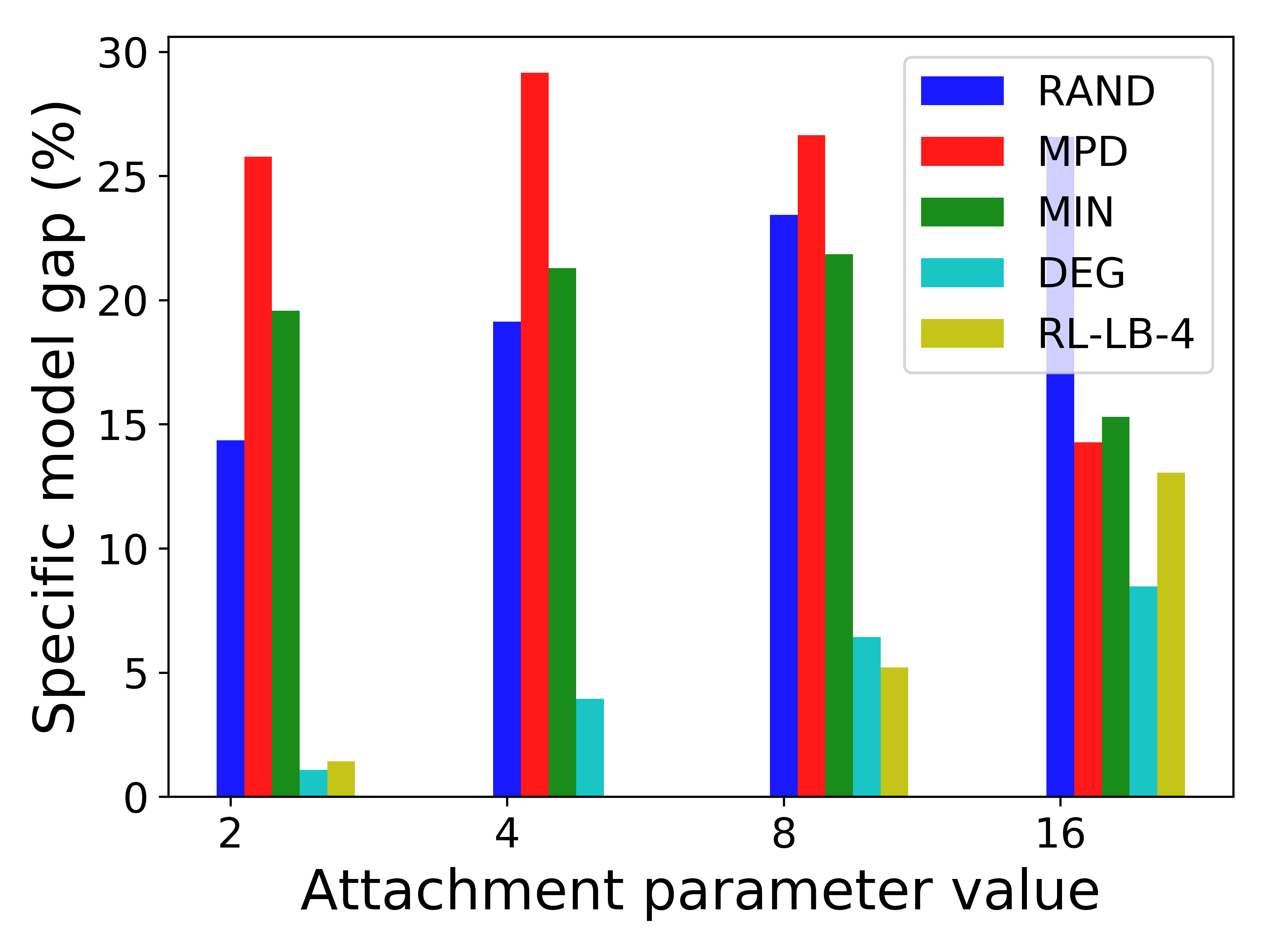}
      \label{sub:enres2}
                         } 
     \caption{Performance on other distributions.}
    \label{fig:node-res-evo}
 \end{figure}

This  set of experiments aim  to analyze the performance of the learned models when they are tested on a  different distribution than  that used for training. 
Figure \ref{fig:node-res-evo} presents the relative gap with the model specifically trained on the distribution tested. For instance, when $\nu=8$, the gap is computed using \texttt{RL-UB-8} as reference (or using \texttt{RL-LB-8} for restricted DDs). We use this measure instead of the optimality gap in order to nullify the impact of the instance difficulty. The gap is then null for the distribution used as reference (\texttt{RL-UB-4} and \texttt{RL-LB-4}). Results show that the more the distribution is distant from the reference, the greater is the gap, which indicates that the learned model strives to generalize. For small perturbations ($\nu=2$ and $8$), good performances are still achieved. These results suggest that it is important to have clues on the distribution of the graphs that we want to access in order to feed appropriately the model during training.

%\begin{figure}[ht]
%  \begin{center}
%    \subfloat[Relaxed DDs ($w = 100$).]{
%      \includegraphics[width=0.49\columnwidth]{results/distorsion-evolution-relaxed}
%      \label{sub:enres1}
%                         }
%    \subfloat[Restricted DDs ($w = 2$).]{
%      \includegraphics[width=0.49\columnwidth]{results/distorsion-evolution-restricted}
%      \label{sub:enres2}
%                         } 
%     \caption{Performance on other distributions.}
%    \label{fig:node-res-evo}
%       \end{center}
%\end{figure}

\subsubsection{Analysis of the Ordering for Relaxed/Restricted DDs}

\begin{figure}[!ht]
  \begin{center}
    \subfloat[Replay of Figure \ref{sub:prel2}.]{
      \includegraphics[width=0.48\columnwidth]{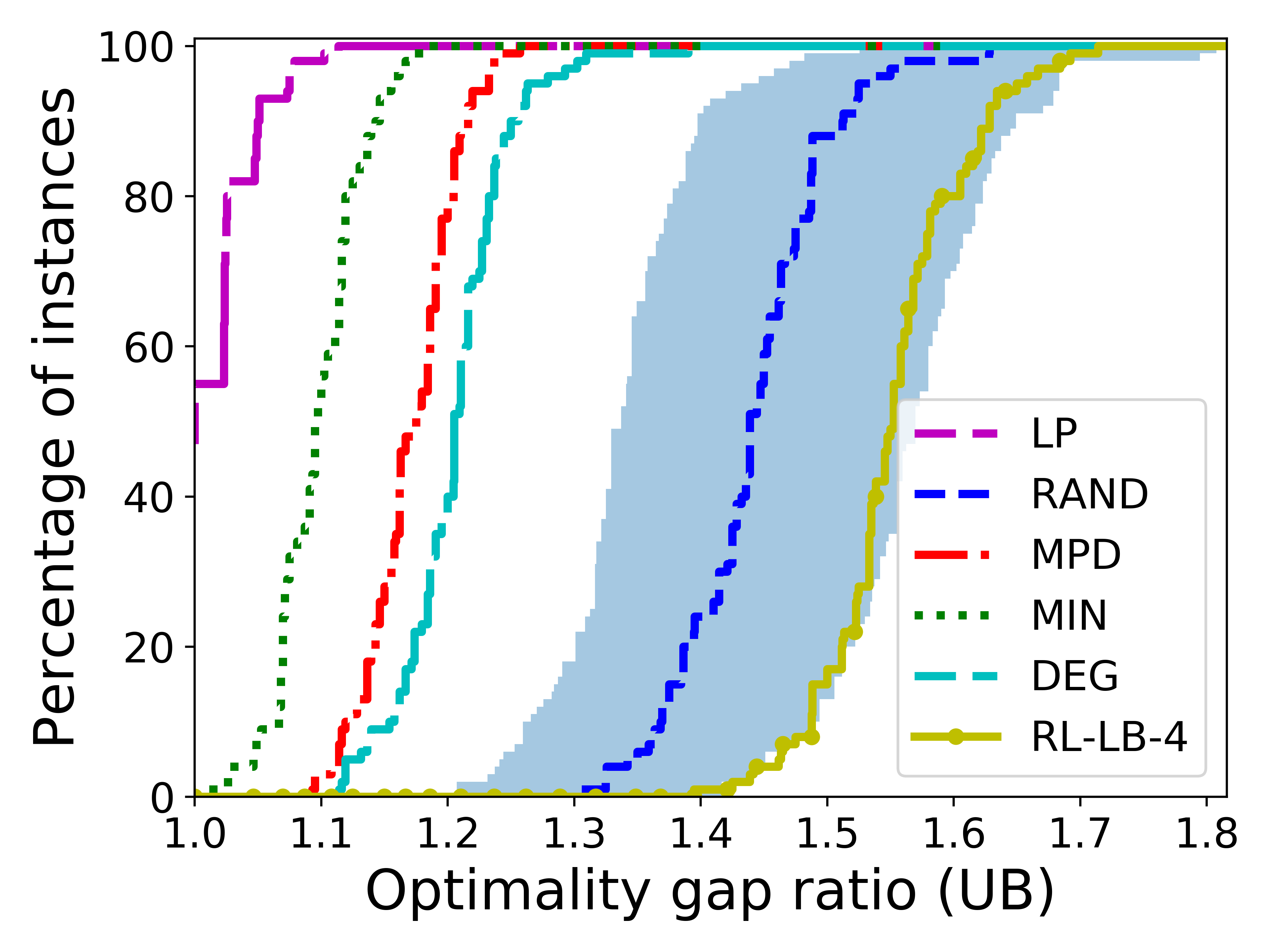}
      \label{sub:other1}
                         }
    \subfloat[Replay of Figure \ref{sub:pres2}.]{
      \includegraphics[width=0.48\columnwidth]{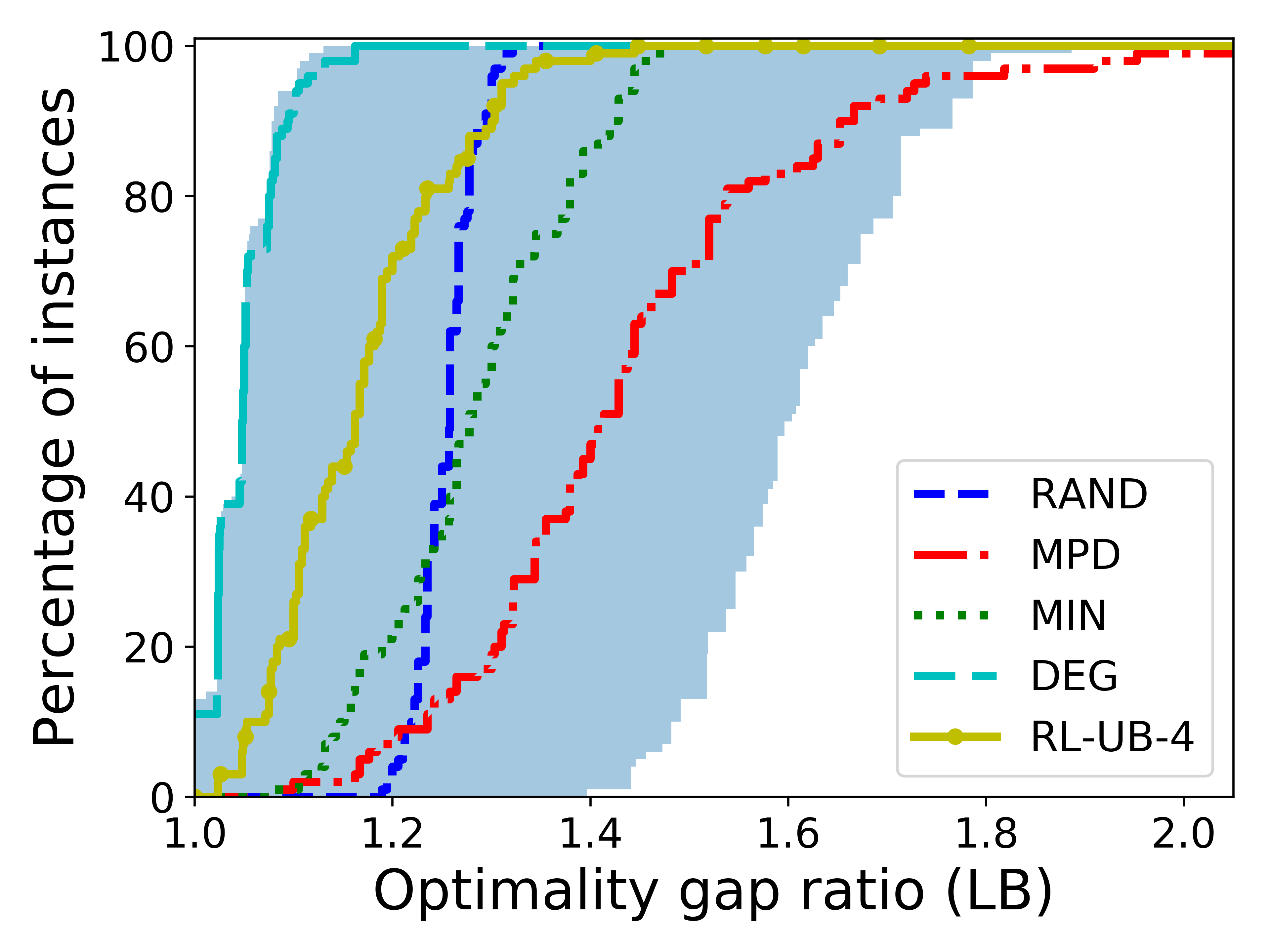}
      \label{sub:other2}
                         } 
     \caption{Performance of the model trained with restricted DDs on relaxed DDs and inversely.}
    \label{fig:node-other-tests}
       \end{center}
\end{figure}

In our initial situation,  we have considered a different model for learning the variable ordering related to relaxed and restricted DDs. Two separate models have then to be trained for getting the bounds of a single instance.
An interesting question that may arise is the following: is a model trained using relaxed DDs can also be used for getting the ordering of restricted DDs or inversely ? The benefit is that only one model would be required for computing the bounds which will reduce consequently the training time. The applicability of this is the purpose of this set of experiments.
Figure \ref{sub:other1} replays the experiments of Figure \ref{sub:prel2} but using the model trained with restricted DDs (\texttt{RL-LB-4}) instead. Similarly,  Figure \ref{sub:other2} shows the situation of  Figure \ref{sub:pres2} with \texttt{RL-UB-4}.  
As we can see, the models do not perform well in such cases and are similar or even worse than the random ordering.
It indicates that both cases must be handled separately. On a higher level of perspective, it empirically shows that an ordering which is efficient for one situation would not be irremediably good for the other one.

\subsubsection{Experiments on the Maximum Cut Problem}

\begin{defi}[Maximum Cut Problem]
\label{def:maxcut}
Let $G(V,E)$ be a simple undirected graph.
A maximum cut of $G$ is a subset of nodes $I\subseteq V$ such that  $\sum_{(u,v) \in C} w(u,v)$ is maximized, where $C \subseteq E$ is the set of edges
having a node in $S$ and the other one in $V \backslash S$. The Maximum Cut Problem (MCP) is that of finding a maximum cut.
\end{defi}

\begin{figure}[!ht]
   \centering
    \subfloat[Relaxed DDs ($w = 100$).]{
      \includegraphics[width=0.48\columnwidth]{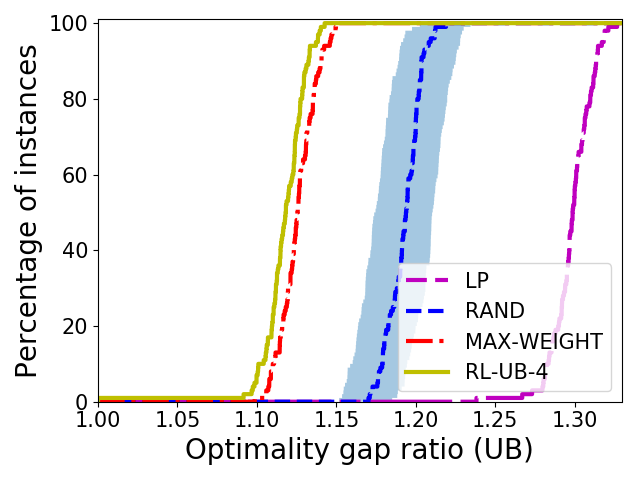}
      \label{sub:mc-prel1}
                         }
    \subfloat[Restricted DDs ($w = 2$).]{
      \includegraphics[width=0.48\columnwidth]{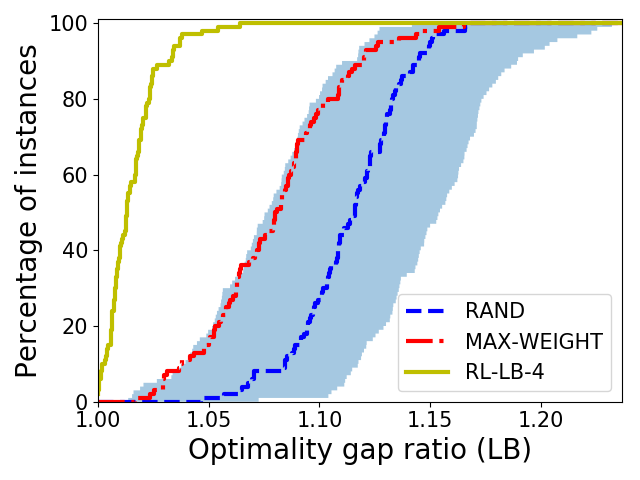}
      \label{sub:mc-pres1}
                         }               
    \caption{Performance profiles for the MCP ($\nu = 4$).}
    \label{fig:maxcut}
\end{figure}

As an example of its generalizability, our approach is also applied to the MCP.
The DD is built according to formulation of \cite{bergman2016decision}. The learning process and the model is the same as for the MISP.
Generation of graphs is still done using a Barabasi-Albert distribution ($\nu=4$) with edge weights uniformly and independently generated from $[1,10]$. 
For training, weights are scaled with a factor of 0.01.  The ordering obtained is compared with \texttt{RAND} and with the \texttt{MAX-WEIGHT} heuristic which selects the vertex having the highest sum of incoming weights \cite{bergman2016decision}. The linear relaxation of a standard integer program of the MCP  \cite{kahruman2007greedy} is also considered. Results reporting the optimality gap of the three methods are presented in Figure \ref{fig:maxcut}  for relaxed ($w=100$ for  training and  testing) and restricted DDs  ($w=2$). 

In both cases, performances better than \texttt{RAND} and \texttt{MAX-WEIGHT} are reached, indicating that the learning is effective.
Concerning relaxed DDs, the gap with \texttt{MAX-WEIGHT} is tighter, which could indicate that this heuristic already gives strong bounds and is then difficult to beat. Finally, the classical linear relaxation does not perform well on the MCP, which was already known \cite{avis2003stronger}.

\section{Conclusion}

Objective function bounds are paramount to general and scalable  algorithms for combinatorial optimization.  Decision diagrams provide a novel and flexible mechanism for obtaining high-quality bounds whose output is amenable to improvement through machine learning, since the objective function bound obtained is directly linked to the heuristic choices taken. This paper provides a generic approach based on deep reinforcement learning for finding  high-quality heuristics for variable orderings
that are shown experimentally to  tighten the bounds proven by approximate DDs. 
Experimental results indicated the promise of the approach when applied to the Maximum Independent Set Problem. 

Insights from a thorough experimental evaluation indicate:
(1) the approach generally outperforms variable ordering heuristics appearing in the literature;
(2) the width chosen during training can have a negligible impact when applied to unseen instances;
(3) the model generalizes well when the width is increased and, in most cases, is applied to larger graphs; 
(4) a separate model must be trained for relaxed and restricted DDs;
(5) the approach generalizes to other problems, such as the Maximum Cut Problem; and
(6) it is important to have a measure of the distribution on the evaluated graphs in order to be able to feed the model during training. 
This last point remains a challenge when extending the approach to  real-world problems.
As a future work, we plan to tackle it by generating new instances for training using generative models from the initial graphs. The idea is to augment the training set by generating new instances, that looks similar in structure to the initial instances, but that are still different.

To the best of our knowledge, this is the first paper to propose the use of machine learning in discrete optimization algorithms for the purpose of learning both primal and dual bounds in a unified framework. 
It opens new insights of research and multiple possibilities of future work, such as the application to different domains that utilize DDs as constraint programming, planning or verification of systems. 
%In a similar fashion, other heuristic choices can be done via reinforcement learning throughout the construction process of approximate DDs in order to improve objective function bounds.

\bibliographystyle{aaai}
 \bibliography{AAAI19-bibli}

\end{document}